\documentclass[preprint,12pt]{elsarticle/elsarticle}
\usepackage{multirow}

\newcommand{\mypara}[1]{\vspace*{0.75ex}\noindent{\bf #1}~}

\newcommand{\be}{\begin{itemize}}  
\newcommand{\ee}{\end{itemize}}  
\newcommand{\bc}{\begin{center}}  
\newcommand{\ec}{\end{center}}  
\newcommand{\bl}{\begin{flushleft}}  
\newcommand{\el}{\end{flushleft}}  
\newcommand{\bq}{\begin{quote}}  
\newcommand{\eq}{\end{quote}}

\newcommand{\bmp}{\begin{minipage}}  
\newcommand{\emp}{\end{minipage}}

\newtheorem{instance}{Example}
\newcommand{\bexa}{\begin{instance}\rm}
\newcommand{\eexa}{\end{instance}}
\newcommand{\eeeexa}{\end{itemize}\end{instance}}
\newcommand{\eeeeexa}{\rule[-0.1mm]{1.0mm}{3mm}\end{itemize}\end{itemize}\end{example}}

\usepackage{figsize}
\usepackage{soul,color}
\usepackage{graphicx}
\usepackage[table]{xcolor}
\usepackage{enumerate}   
\usepackage[ruled,vlined,noend]{algorithm2e}
\usepackage{amsmath, amssymb}
\usepackage{tikz}
\tikzset{every picture/.style={/utils/exec={\sffamily}}}
\usetikzlibrary{positioning}
\usepackage{textcomp}
\usepackage{xcolor}
\usepackage{mathtools}
\usepackage{hyperref}

\DeclarePairedDelimiterX{\infdivx}[2]{[}{]}{
  #1\;\delimsize\|\;#2%
}

\newcommand{\pure}{\textsc{PURE}}
\newcommand{\peak}{\textsc{PEAK}}

\AtBeginDocument{%
  \providecommand\BibTeX{{%
    \normalfont B\kern-0.5em{\scshape i\kern-0.25em b}\kern-0.8em\TeX}}}

\makeatletter
\renewcommand{\SetKwInOut}[2]{%
  \sbox\algocf@inoutbox{\KwSty{#2}\algocf@typo:}%
  \expandafter\ifx\csname InOutSizeDefined\endcsname\relax
    \newcommand\InOutSizeDefined{}%
    \sbox\algocf@inoutbox{\KwSty{#2}\algocf@typo\textbf{:}~}\setlength{\inoutindent}{\wd\algocf@inoutbox}%
  \else
    \ifdim\wd\algocf@inoutbox>\inoutsize%
    \sbox\algocf@inoutbox{\KwSty{#2}\algocf@typo\textbf{:}~}\setlength{\inoutindent}{\wd\algocf@inoutbox}%
    \fi%
  \fi
  \algocf@newcommand{#1}[1]{%
    \ifthenelse{\boolean{algocf@inoutnumbered}}{\relax}{\everypar={\relax}}%
    {\let\\\algocf@newinout\hangindent=\inoutindent\hangafter=1\KwSty{#2}\algocf@typo\textbf{:}~##1\par}%
    \algocf@linesnumbered
  }}%
\makeatother

\begin{document}

\title{\peak: Explainable Privacy Assistant through Automated Knowledge Extraction}

\author[1]{Gonul Ayci\corref{cor1}}
\ead{gonul.ayci@boun.edu.tr}

\author[2]{Arzucan \"Ozg\"ur}
\ead{arzucan.ozgur@boun.edu.tr}

\author[3]{Murat \c{S}ensoy}
\ead{drmuratsensoy@gmail.com}

\author[4]{P{\i}nar Yolum}
\ead{p.yolum@uu.nl}

\cortext[cor1]{Corresponding author}   

\affiliation[1]{organization={Computer Engineering, Bogazici University}, 
             country={Turkey}}

\affiliation[2]{organization={Computer Engineering, Bogazici University}, 
            country={Turkey}}

\affiliation[3]{organization={Amazon Alexa AI}, 
                country={UK}}

\affiliation[4]{organization={Information and Computing Sciences, Utrecht University}, 
                country={The Netherlands}}

\begin{abstract}
In the realm of online privacy, privacy assistants play a pivotal role in empowering users to manage their privacy effectively.~Although recent studies have shown promising progress in tackling tasks such as privacy violation detection and personalized privacy recommendations, a crucial aspect for widespread user adoption is the capability of these systems to provide explanations for their decision-making processes.~This paper presents a privacy assistant for generating explanations for privacy decisions.~The privacy assistant focuses on discovering latent topics, identifying explanation categories, establishing explanation schemes, and generating automated explanations.~The generated explanations can be used by users to understand the recommendations of the privacy assistant.~Our user study of real-world privacy dataset of images shows that users find the generated explanations useful and easy to understand.~Additionally, the generated explanations can be used by privacy assistants themselves to improve their decision-making.~We show how this can be realized by incorporating the generated explanations into a state-of-the-art privacy assistant.
\end{abstract}


\maketitle

\section{Introduction}
\label{intro}
Many of the computing systems we use today are facilitating and encouraging the sharing of information on a large scale.~Billions of users on online social networks (OSNs) are exchanging images, videos, and comments with each other constantly.~At the same time, we rely on the Internet of Things (IoT) applications, such as smart home systems, for many everyday activities, which in return collect, store, and fuse our everyday data to provide their services.~On one hand, these systems have become extremely undeniably beneficial, but on the other hand, people are finding it challenging to manage their privacy while using them. 

We understand privacy management as the task of handling who has access to personal information, for what purposes, and in which context. 
It includes protecting users from being adversely affected by information shared about them, for example, information shared without their consent or awareness on OSNs.~Interestingly, the information that can violate a user's privacy can come from different sources and in different forms.~It could be that users post information about themselves that later causes them harm, such as losing their job or being investigated due to controversial comments~\cite{nissenbaum2004privacy, nissenbaum2009privacy}.~Additionally, others might post content about the user on the user's profile page without their knowledge or consent, such as tagging identifiable, geotagged images~\cite{kokciyan2016p}.

The use of systems such as OSNs and IoT applications has become pervasive due to the provision of useful services, such as document sharing and home entertainment, but users are increasingly concerned about their privacy and often self-censor or delete content after sharing~\cite{cook2013two}.~The difficulty in managing privacy is compounded by the discrepancy between users' preferences and their actual behavior~\cite{iachello2007end}, and the fact users must constantly decide whether they want to share content or not.~Due to the vast amount of content, users often do not have the time to make informed decisions, resulting in decision fatigue and making them prone to errors.~Additionally, various recent surveys conducted with users of online social networks indicate that people do not even read the privacy policies that they accept~\cite{vila2004we,acquisti2005privacy}.

Privacy assistants can collaborate with users, offering valuable support in the management of their privacy-related responsibilities~\cite{fogues2017sosharp,colnago-2020}.~When users engage in content sharing, thoughtful consideration regarding the intended recipients and the required configuration of privacy settings becomes essential~\cite{ulusoy2021panola}.~Images are an important category amongst the various types of content shared online. Recent research has focused on enabling users to classify whether an image should be regarded as private or public~\cite{tonge2020image}.~This classification can prove beneficial in preventing the unintentional sharing of private images on OSNs.~Given the personal and subjective nature of privacy, where users’ perceptions of privacy may vary, it becomes imperative for an assistant to deliver personalized responses concerning the privacy status of images~\cite{kurtan2021assisting,ayci2023uncertainty}.~To ensure the widespread acceptance of privacy assistants among end users, establishing trust becomes crucial.~One effective approach to cultivate such trust is through the provision of explanations~\cite{mosca2022explainable}.~Consequently, we tackle a novel challenge related to the interplay between explanations and privacy: how can a privacy assistant explain the rationale behind classifying a particular piece of content as either private or public?

This paper presents a privacy assistant~\peak~that decides whether a given image is private or not, and explains this decision to its user as well as other privacy assistants. To the best of our knowledge, this is the first privacy assistant that can explain why an image is considered private or public.

Explainable Artificial Intelligence (XAI) suggests approaches that aid in the comprehension of why and how a machine learning (ML) model arrives at its prediction.~There are various explanation methods with respect to visual and textual explanations~\cite{simonyan2013deep, lundberg2017unified, lundberg2020local2global, wachter2017counterfactual}.~For the image privacy prediction task, an example of a visual explanation can be highlighting the most important region in the image for the target class, whereas a textual explanation can be a generated text such as ``if a guitar had not been in the image, the image would not have been public"~\cite{wachter2017counterfactual}.~A visual explanation can quickly point out a potential privacy concern, however, it can be overwhelming for users to understand underlying mechanisms since it may present many intricate details~\cite{simonyan2013deep, lundberg2017unified, adebayo2018sanity, ghorbani2019interpretation, kindermans2019reliability}.~Szymanski~{\it et al.}~\cite{szymanski2021visual} suggests that the combination of visual and textual explanations improves understanding for non-expert (without any software or domain knowledge) users. We propose an explanation method that provides a visual representation accompanied by a textual description. 

\peak~utilizes automatically generated tags (descriptive keywords) of images, explores latent topics from the tag sets, classifies images based on image-topic associations, and ultimately generates human-understandable explanations.~An explanation involves presenting one or more topics related to an image in a visual format, highlighting significant tags associating the image to a certain topic, and providing a textual description detailing the connections between the topics.~\peak~derives these explanations from a well-known image dataset for privacy where images are labeled as public or private~\cite{zerr2012privacy}.~We perform a user study to gauge if participants actually find the generated explanations useful and what factors of the explanation or the image affect users' understanding of the model decision.~Furthermore, privacy assistants may delegate some privacy decisions to their users to avoid failures when they are uncertain. This generally improves the accuracy of privacy decisions~\cite{ayci2023uncertainty} but leads to increased cognitive overload for the user.~An important benefit of \peak~is that it enables privacy assistants to base their decisions on explanations, resulting in fewer images to be delegated to users, thus reducing the cognitive overload for users without compromising the accuracy of privacy decisions.~Minimizing extraneous cognitive load is important in order to ensure sufficient cognitive resources are available for the remaining delegated images~\cite{skulmowski2022understanding}.~Since the cognitive load of users increases with the complexity of explanations and the number of topics, we are trading off the comprehensibility for the expressiveness of the explanations in this work.

The rest of this paper is organized as follows.~Section~\ref{sec-relatedwork} discusses our work in relation to related work. Section~\ref{sec-system} explains the mechanism of our privacy assistant in detail.~Section~\ref{sec-extracting-features} develops our privacy assistant into a system that can be readily used to explain the privacy labels of images.~Section~\ref{sec-generateexp} presents how to identify explanation categories and generate explanations.~Section~\ref{sec-userstudy} evaluates our explanation model with a user study.~Section~\ref{sec:result_performance} shows experimental results of \peak~working in combination with privacy assistants.~Finally, Section~\ref{sec-conclusion} concludes our work and outlines future directions.

\section{Related Work}\label{sec-relatedwork}
There are two bodies of research that are important for this work. The first one is work on predicting whether an image is private or not (Section~\ref{sec:predict}) and the second one is work on explainability techniques (Section~\ref{sec:xai}).

\subsection{Image Privacy Prediction Approaches}
\label{sec:predict}
In the literature, several studies on image privacy prediction make use of descriptive keywords (tags) and visual features. Squicciarini {\it et al.} \cite{squicciarini2017tag} present a Tag-To-Protect (T2P) system that automatically recommends privacy policies based on the image tags. Their study shows that prediction accuracy decreases when there are large tag sets and when the number of tags per image increases. Tonge and Caragea \cite{tonge2020image} use deep visual semantics (i.e.~deep tags) and textual features (i.e.~user tags) to develop a model that predicts whether an image is private or public. Deep tags of images are the top \textit{k} predicted object categories extracted from pre-trained models. Using user-created tags, they create deep visual features by adding highly correlated tags to visual features extracted from the fully connected layer of the pre-trained models. They use Support Vector Machine (SVM) classifiers with pre-trained CNN architectures such as AlexNet, GoogLeNet, VGG-16, and ResNet to extract features (tags). They find that a combination of user tags and deep tags from ResNet with the top $350$ correlated tags performs well for privacy prediction, as adding the highly correlated tags improves prediction performance. Kurtan and Yolum \cite{kurtan2021assisting} propose an agent-based approach, namely PELTE, which addresses the same problem with automatically generated image tags. The internal tag table stores the data of privacy labels collected from images shared by the user. The external tag table stores the data of images shared by the user's friends. Their proposed system performs well in predicting privacy, even though the personal assistant only has access to a small amount of data. Ayci {\it et al.} \cite{ayci2023uncertainty} propose a personal privacy assistant called PURE to preserve the privacy of its user. PURE is aware of uncertainty by generating an uncertainty value for each prediction of a given image, informing its user about it, and delegating decisions back to the user if it is uncertain about its predictions. PURE is able to make personalized predictions by using the personal data of its user. It is also risk-averse, by incorporating the user's risk of misclassification. Their experiments are fruitful in analyzing the link between uncertainty and misclassification. They show that PURE captures uncertainty well and performs better compared to alternative models for quantifying uncertainty (i.e.~Monte Carlo dropout \cite{gal2016dropout} and Deep Ensemble \cite{lakshminarayanan2017simple}). Although they demonstrate the success of using descriptive keywords and visual features to predict image privacy, neither of these approaches addresses capturing the explanations for the privacy predictions as we have done here. However, explaining the model predictions is critical to understanding people's privacy expectations and preferences. In this study, we propose a novel methodology that uses descriptive keywords to explore latent topics by topic modelling and provides explanation schemes for predictions.

An important body of work uses taxonomies in classification. A taxonomy categorizes content and user preferences based on certain characteristics and attributes.~It identifies important features and patterns in content.~Orekondy {\it et al.} \cite{orekondy2017towards} present a model for the privacy risk prediction task for images and provide $68$ privacy attributes such as \textit{nudity, passport} and \textit{religion}.~Li {\it et al.}~\cite{li2020towards} propose a method to find out what kind of visual content is private.~They develop a taxonomy with $28$ categories such as \textit{nudity/sexual, irresponsible to child} and \textit{bad characters/unlawful/criminal}.~Zhao~{\it et al}~\cite{zhao2022privacyalert} define a privacy taxonomy with $10$ categories with the most commonly used descriptive keywords for a certain category.~For example, the descriptive keywords of the category \textit{religion/culture} include \textit{culture, religion, and spiritual}.~Even though these studies propose inspiring taxonomies for privacy, their approaches do not provide explanations for a particular image as to why the image is labeled public or private, as we have done here. 

\subsection{Explanation approaches}
\label{sec:xai}
Gilpin~{\it et al.}~\cite{gilpin2018explaining} provide a broad perspective of explainable AI (XAI) systems and identify three categories of approaches for providing explanations in ML systems: processing, representation, and explanation-producing. Processing involves emulating the data processing of an ML system to establish connections between input and output in order to justify emitted choices.~Representation refers to how the internal data structures and operations of the network can be explained to gain insight into why confident choices are made.~Explanation-producing refers to generating human-understandable descriptions of the model's behavior and decision-making process.~They suggest evaluating explanation-producing models by how well they align with human expectations. The role of human evaluation in interpreting ML models is significant as it allows for assessing the reasonableness of the explanations provided by the models. Riveiro and Thill~\cite{riveiro2021s} discuss the need for human evaluations in the field of XAI, specifically about how users interact with systems that generate explanations of AI systems.~They highlight a gap between user expectations and the explanations provided by the system and suggest the need for more user-centered approaches to designing explanations.~A factual explanation provides information about the reasons why a certain output was produced by an AI system, whereas a counterfactual explanation suggests alternative outcomes that could have occurred if the input to the system had been different.~They find that factual explanations are appropriate when the system output aligns with user expectations.~However, when there is a mismatch between expectations and output, neither factual nor counterfactual explanations are appropriate and counterfactual explanations may not be sufficient on their own.~The results also suggest that the accuracy of the user's mental model of the AI system is connected to the effectiveness of the explanations provided and that further exploration of the context and details of the global system model may be useful in this regard.~Langley~\cite{langley2019explainable} explores the concept of explainable agency and its relationship with a normative and justified agency.~The paper emphasizes the importance of creating intelligent agents that are capable of explaining their actions to humans in a way that aligns with human values and norms.~They suggest that this can be achieved by designing agents that are able to reason about their own decisions and the reasons behind them, and by providing explanations that are understandable and relevant to humans.~The paper proposes a framework for developing such agents, which involves incorporating normative principles into the agent's decision-making process and using explanations to justify the agent's actions.~They highlight the importance of creating agents that are transparent, understandable, and aligned with human values in order to ensure that humans can trust and rely on them.~Miller~\cite{miller2019explanation} examines studies of explainability within the scope of philosophy, social and cognitive psychology, and cognitive science.~Their study provides various definitions of explainability, criteria for selecting explanations, evaluating explanations, and useful insights for XAI.~They define the interpretability of a model as the degree to which the cause of a prediction can be understood.~In the context of AI, generating an explanation that establishes a shared understanding of the decision-making process between an intelligent agent and a human observer is crucial.~Justification is provided by explaining why a decision is good.~Arrieta~{\it et al.}~\cite{arrieta2020explainable} provide an overview of XAI by defining interpretability and explainability.~They define interpretability as the ability to explain meaning in a form that people can understand.~They associate explainability with an explanation as the interface between a human and a decision-maker. 

\mypara{Interpretable Models:}These models enhance the interpretability and transparency of their decision-making processes. Linear Regression, Logistic Regression, Decision Trees, Random Forest, and k-nearest neighbors provide interpretable insights into their predictions. For instance, the Random Forest model can be trained to predict whether a given image is public or private. It enables us to trace and visualize the decision paths to understand which features of the image had a major impact on privacy predictions. While this is important, since there can be thousands of features, it is not always possible to understand the explanations. 

\mypara{Model-specific Methods:}These explanation methods improve our understanding of how deep neural networks make decisions. The most popular model-specific techniques include Saliency Map and Attention Map. Simonyan {\it et al.} \cite{simonyan2013deep} propose methods for generating visualizations of Convolutional Neural Networks (CNN) classification models trained on the ImageNet dataset by utilizing numerical optimization of the input image. Even though their proposed saliency map of a given image helps to identify the regions of the image that are most discriminative with respect to the given class, it can be limited in capturing all the important information that flows through a deep neural network. Selvaraju {\it et al.} \cite{selvaraju2017grad} present Gradient-weighted Class Activation Mapping (Grad-CAM) method that generates visual explanations for any CNN, regardless of its architecture. As a visual explanation for a given image, Grad-CAM provides a heatmap that would have the greatest impact on the output. Xiao {\it et al.} \cite{Xiao_2015_CVPR} propose a two-level attention model for fine-grained classification tasks that can highlight discriminative regions at different levels of abstraction in an image. The first level of attention involves object-level filtering that selects relevant patches to feed into the classifier. The second level of attention conducts part-level detection to detect the parts of the classification. Since these methods are highly dependent on their architectures, their explanations cannot generalize well to different architectures (i.e.~not CNN). Additionally, the units of attention-based approaches are not explicitly trained to provide human-understandable explanations. They can be also vulnerable to adversarial examples, which have the potential to deceive a model and flip the prediction (i.e.~misclassification) by doing so.

\mypara{Model-agnostic Methods:}These explanation techniques can be applied to any ML model, regardless of its architecture. Model-agnostic methods for the binary classification task consider what features of the input have been influential on the decision. Ribeiro {\it et al.} propose \cite{ribeiro2016should} the local interpretable model-agnostic explanation model (LIME) that generates a new dataset consisting of perturbed samples and the corresponding predictions of the black box model. It then trains an interpretable model (e.g., Random Forest) on this new dataset. Lundberg {\it et al.} \cite{lundberg2017unified} propose a model-agnostic feature relevance explanation model (SHAP), that is based on a game theoretically Shapley values \cite{shapley1997value}. This method computes the contribution of each feature to the prediction output. Slack {\it et al.} \cite{slack2020fooling} show that LIME is more prone to being deceived compared to SHAP. LIME also may not be an appropriate method for explaining image privacy predictions, as it may not be able to effectively handle the complex relationships between features, where small changes can have a big impact on the model's predictions. Furthermore, the method of perturbing the input used by LIME can be easily recognized by attackers. On the other hand, SHAP explanations are based on Shapley values, which can be difficult to interpret by non-expert users due to the highlighting of the contributions of each feature. In addition, explanations may be more difficult to understand when SHAP has a high-dimensional feature space.

\mypara{Example-based Explanations:}These explanations utilize specific examples from the dataset to understand how minor changes in a given image would affect the model's prediction and how the model made the prediction for the image.~Example-based explanations provide examples from the dataset instead of displaying feature importance.~Wachter {\it et al.}~\cite{wachter2017counterfactual} propose counterfactuals explanations that are a way of understanding the causal relationship between inputs and the model's predictions by simulating hypothetical scenarios and analyzing how minor changes to the feature of a given image affect the prediction.~The conventional notion of ``explanation" in the AI literature involves explaining the internal state or logic of an algorithm that leads to a decision, whereas counterfactual explanations focus on the external factors that led to that decision.~This is a critical distinction since ML algorithms can have millions of interrelated variables, making it difficult to convey their behavior to non-expert users.~Counterfactual explanations help to interpret the predictions; however, multiple counterfactual explanations can be generated for a given image.~It can be challenging to decide which one to use in this case.

Despite the fact that these methods can assist in providing explanations for the model's predictions, they can also encounter difficulties that can make them unable to produce human-understandable explanations. For example, explanations can include too many features or they may not be representative of the model's behavior. 

An important work of explainability in conjunction with privacy, is by Mosca and Such~\cite{mosca2022explainable}, where they develop an agent that uses computational argumentation to resolve disputes and propose a text-based description of the outcome of the argumentation, generated by the system. They suggest a framework for generating explanations consisting of two types of explanations: general and contrastive. A general explanation provides an overview of the agent's decision-making process, without providing specific details about the reasons behind a particular decision. They are useful for giving users an understanding of how the system works and what factors it considers when making decisions. On the other hand, a contrastive explanation provides specific reasons for a particular decision and is intended to help users understand how they might modify their behavior to achieve better outcomes.~These types of explanations are particularly useful for situations where the agent's decision might conflict with a user's goals or preferences.~While explaining how the system operates is useful, that work does not provide an explanation as to why a given piece of content is private or public.  Our focus, on the other hand, is on explaining to both end-users and personal assistants why a given image can be considered public or private. \peak~is a model-agnostic approach that can be applied to any ML model.

\section{\peak: Explainable Privacy Assistant}\label{sec-system}
\begin{figure}[htb]
    \centering
    \includegraphics[scale=0.5]{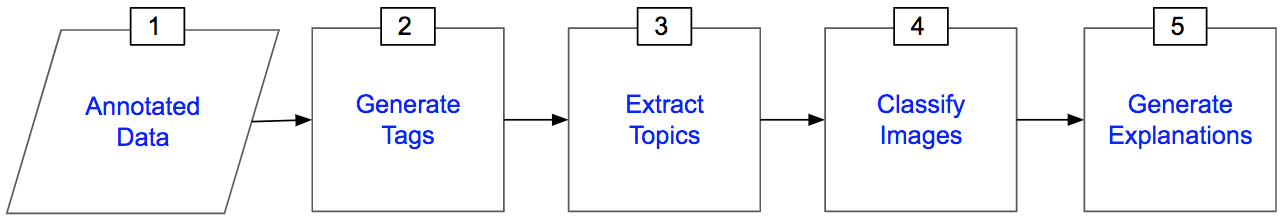}
    \caption{\textit{Architecture of \peak}} 
    \label{fig:core_system}
\end{figure}

\peak~is a privacy assistant that explains privacy labels by generating human-understandable explanations.~The generated explanations revolve around topics (collections of related keywords) as these are intuitive and easier to understand for users.~An image can belong to multiple topics with varying degrees of relatedness, and these relationships can help understand why an image is considered to be public or private.~For end users, the most important aspects of an explanation are simplicity and relevance.~Hence, the explanation is visual in nature and accompanied by a short explanatory text describing the connections between the topics and the image.~The explanation is focused on the most important features of the image and associated topics, rather than being an exhaustive analysis.~In order to achieve this, only a subset of the most relevant topics associated with an image are used to explain why the image is classified as public or private.~Ultimately, \peak~helps OSN users by allowing them to understand why images have been identified as private or public.~Moreover, \peak~can aid privacy assistants in enhancing their performance on making privacy decisions (Section~\ref{sec:result_performance}).~Figure \ref{fig:core_system} shows the architecture of \peak.  
%

\peak~is composed of the following five stages: 
\begin{enumerate}
    \item The starting point for \peak~is a set of labeled public and private images.~The data could come from a user's own history of personal online images (i.e.~both shared and not shared), or from a publicly available dataset of labeled images.
    \item The second stage involves assigning tags, i.e.~descriptive keywords, to each image.~The generated tags are in plain language such as ``tree" or ``baby", and they can be provided by users themselves or generated automatically by a tool such as \textit{Clarifai}.
    \item The next step is using the set of labeled and tagged images to perform topic modelling, which is a technique used to extract latent (i.e.~hidden) topics from textual information (i.e.~the tags).~Each image is associated with one or more topics and topics are constructed using two criteria.~First, tags within a single topic should be semantically related, meaning images associated with the same topic should be described by similar tags.~Second, topics should be semantically distinct from each other. In other words, there should not be too much overlap between different topics.~Finally, in order to make it easier to understand for the user, we named the generated topics manually.~For example, the topic with the tags ``tree", ``parks", and ``outdoors" was named Nature.
    \item Once topics have been generated, the next stage involves training a tree-based ML algorithm for binary image classification.~In this case, the generated topics serve as features of the images and the algorithm uses image-topic relationships to predict whether a new image should be classified as public or private. Additionally, the contributions of each topic  to the privacy decision (i.e.~positive or negative effect) are computed. 

    \item The final step is identifying explanation categories for images, which essentially are image profiles with certain characteristics in terms of topic contributions.~For instance, there may be a single dominant topic that pushes the prediction overwhelmingly in one direction.~Based on the topics' contributions to privacy prediction, images are assigned to one of four explanation categories we identified: Dominant, Opposing, Collaborative, and Weak.~Finally, for each explanation category, there is a distinct textual and visual explanation pattern based on the image's topics and the relationships between topics.
\end{enumerate}

For end users, it is important that any explanation generated by \peak~is simple and easy to understand, i.e.~no technical knowledge should be required.~Ultimately, our aim is not to explain the technical process of the classifier to the end user. Additionally, since many users do not actually read long, complicated privacy policies, the explanation should be visually understandable and supported by a short text.

Based on these constraints, we propose an explanation as to why an image is considered public or private using a set of {\it topics} belonging to the image.~Each image can have one or more topics and these topics are shown as a circle which is labeled with the topic name.~Additionally, we identify one or more \textit{tags} linking this image to each topic and denote them in the corresponding topic circle.~The visual representation aims to explain that the image is public or private because it is described by the displayed topics and tags.~Furthermore, the visualization is augmented with a short explanatory description using a predetermined language structure.~The text is thus supplementary and does not provide additional information.
%
\begin{figure}[htb]
  \centering
    \subfigure[Image]{\includegraphics[scale=0.46]{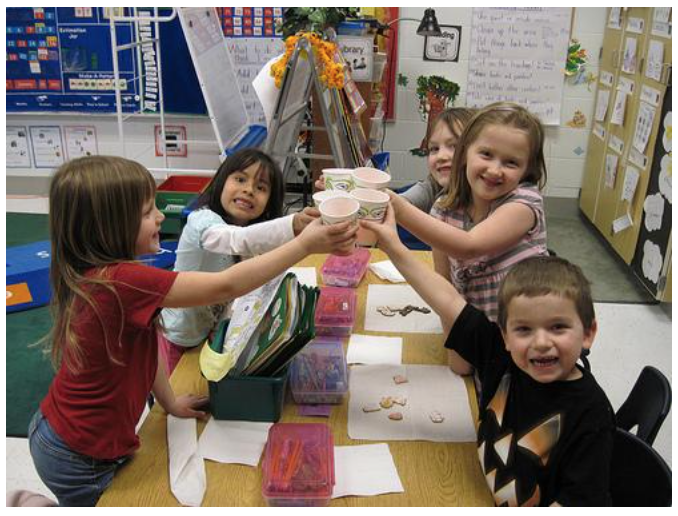} \label{fig:image_example}}
    \quad
    \subfigure[Generated Explanation]{\includegraphics[scale=0.35]{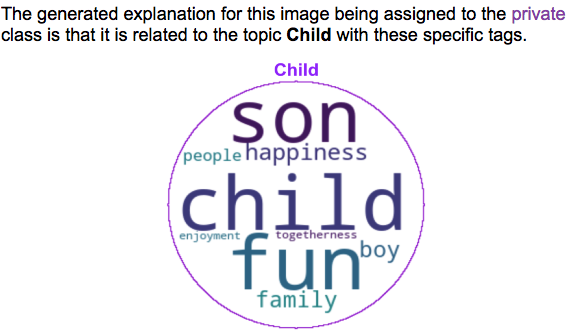}\label{fig:explanation_example}}
  \caption{An example image annotated as private and its generated explanation by \peak}
  \label{fig:image_and_generated_explanation}
\end{figure}

Figure \ref{fig:image_example} displays an example of a private image. Figure \ref{fig:explanation_example} presents the  explanation generated by the \peak~method.~The explanation shows the image is classified as private because it is associated with topic \textit{Child} and its related tags.~Whilst in this example only one topic is displayed, there can be multiple topics contributing to the prediction.~In that case, \peak~visually represents only the most relevant topics and explains the relation between these different topics.

\section{Establishing Topics}\label{sec-extracting-features}
We start walking through the steps outlined in Figure~\ref{fig:core_system}.
\subsection{Generating Tags}
\label{sec:tags}
To obtain our dataset of images (Step $1$), we select a well-balanced subset from the publicly available PicAlert dataset~\cite{zerr2012privacy}, which is widely used for addressing the privacy prediction task in images~\cite{squicciarini2017tag,tonge2020image,kurtan2021assisting,ayci2023uncertainty}. The PicAlert dataset consists of Flickr images that have been labeled as either \textit{public} or \textit{private} by a group of annotators. These annotations were provided by $81$ users spanning an age range of $10$ to $59$ years and representing diverse backgrounds. We consider an image as private if it has been labeled as such by at least one annotator, and as public if all annotators have labeled it as public. The subset we have chosen for our study comprises $32,000$ images, consisting of $27,000$ Training images and $5,000$ Test images. In the subsequent stage of our process, we utilize the Clarifai API\footnote{https://clarifai.com/clarifai/main/models/general-image-recognition} in order to automatically generate a set of $20$ different tags for each image. The architecture of the general image recognition model of Clarifai is convolutional neural networks such as InceptionV2. Their model is trained on over $20$ million images and uses $10,000$ concepts to identify objects in images and videos.
%
\begin{figure}[h]
	\centering
    \subfigure[Image]{\includegraphics[scale=0.46]{figures/image_example.png} \label{fig:img_example}}
    \quad
    \subfigure[Generated Tags]{\includegraphics[scale=0.35]{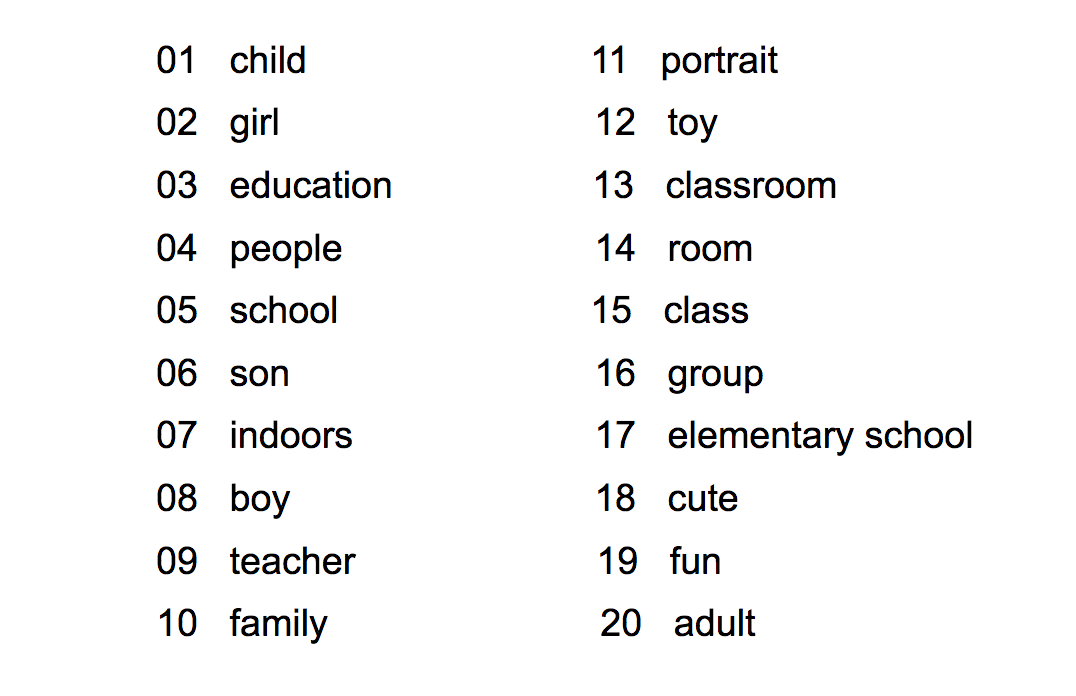}\label{fig:tag_example}}	
    \caption{Example image and its generated tags by Clarifai}
    \label{fig:clarifai_example}
\end{figure}

Figure~\ref{fig:img_example} displays an example image and Figure~\ref{fig:tag_example} shows $20$ different tags generated by Clarifai.~A qualitative assessment of the example image shows it is captured well by the generated tags. 

\subsection{Extracting Topics}\label{sec-generating-topics}
\begin{figure}[h]
	\centering
	\includegraphics[scale=0.4]{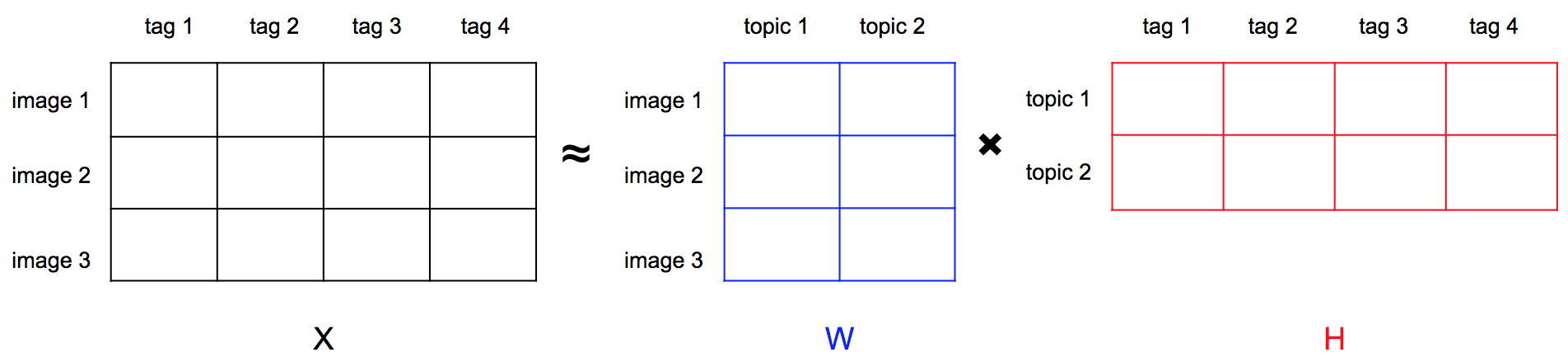}
	\caption{Non-negative Matrix Factorization (NMF) concept}
	\label{fig:nmf_concept}
\end{figure}

Topic modelling is a technique that discovers latent topics within a collection of textual information.~It allows us to extract distinct topics from an image's generated tags.~We employ the widely used Non-negative Matrix Factorization (NMF) topic modelling technique \cite{lee1999learning} (Step $3$).~NMF is an approximation to factorize a non-negative image-tag matrix $X$ into non-negative matrices $W$ and $H$ as illustrated in Figure~\ref{fig:nmf_concept}.~The $W$ (features) matrix denotes the degree to which an image belongs to a topic and the $H$ (components) matrix denotes the degree to which a tag belongs to each topic.~The $W$ and $H$ matrices are initialized randomly.~The NMF algorithm runs iteratively until it finds W and H matrices that minimize the Frobenius norm of the matrix, that is, $\left \| X - W \times H \right \|_{F}$.~NMF is suitable for interpretability (components are non-negative) and works better and faster for short texts (a set of tags) compared to alternatives such as Latent Dirichlet Allocation (LDA) \cite{blei2003latent}.~In this study, we make use of the term weighting method and we specifically employ the  Term Frequency - Inverse Document Frequency (TF-IDF) model to measure the presence of tags.~TF-IDF allows us to transform tags into numerical vectors in order to construct an \textit{image-tag (X)} matrix.~We learn the tag vocabulary from the images and then transform the images into a \textit{image-tag (X)} TF-IDF matrix.~Each row of the matrix corresponds to an image, and each column represents a unique tag from the images.~The value in each cell of the matrix represents the TF-IDF weight of the tag in the corresponding image.~We build the NMF model for a given number of topic (k) values, which generates an \textit{image-topic (W)} matrix and a \textit{topic-tag (H)} matrix.~Note that the number of topics $k$ is an important parameter to set, which we explain next.
%

\subsection{Evaluation of Topics}
Topics should be meaningful and interpretable for humans.~One way of realizing this is to ensure that the topics are {\it coherent}, which means topics should be relatively different from each other (i.e.~distinct) whilst images within a certain topic should be described by similar keywords.~Hence, we can measure coherence based on two different criteria as follows:

\begin{enumerate}[(i)]
\item Intra-topic similarity: The average semantic similarity between all pairs of the most associated \textit{N} tags in the same topic.~Hence, it is a measure of how similar a topic's tags are to each other.~For instance, the most associated three tags with \textit{Topic Nature} and \textit{Topic Child} are $\left \{tree, park, wood\right \}$ and $\left \{child, baby, fun\right \}$, respectively.~In our example, both topics have high intra-topic similarity.

\item Inter-topic similarity: The average semantic similarity of the most associated \textit{N} tags from different topics. Hence, it is a measure of how much overlap there is between different topics' tags. For example, the associated tags with \textit{Topic Nature} and \textit{Topic Child} are clearly not overlapping, meaning they have a low inter-topic similarity.
\end{enumerate}

We can evaluate the quality of the explored topics by calculating these two measures of similarity. For good topic modelling, we want to 1) maximize the intra-topic similarity, thus ensuring the topic is well-defined and the associated tags are closely related to each other, and 2) minimize inter-topic similarity, thereby ensuring the topics are distinct and the tags within each topic are not closely related to the tags in the other topics.

In the NMF model, we set the number of topics based on the model's performance in terms of \textit{coherence}. While calculating intra-topic and inter-topic similarity, each tag is represented by word embedding vectors, namely word2vec \cite{mikolov2013distributed}. We represent tags as $300-$dimensional vectors of the word2vec model trained on Google News when calculating coherence\footnote{https://code.google.com/archive/p/word2vec/}. The similarity between two tag vectors is measured by the \textit{cosine similarity} metric. Semantically similar tags tend to be close to each other in the semantic space. Intra-topic similarity values for $10$ and $20$ topics are $0.18$ and $0.20$, respectively. Note that the cosine-similarity values between two vectors for this model are generally low. For instance, the similarity between ``person" and ``people" is $0.51$ and ``tree" and ``park" is $0.23$. Hence, $k=20$ is better in terms of intra-topic similarity as the tags within a given topic are more related to each other. Additionally, inter-topic similarity values for $10$ and $20$ topics are $0.48$ and $0.43$, respectively. In this case, $k=20$ is again a better fit as the lower value indicates more distinct, i.e.~better segregated, topics.
Thus, we set the number of topics to $20$.

%

\begin{figure*}
\centering
\SetFigLayout{1}{5}
  \subfigure[Nature]{\includegraphics{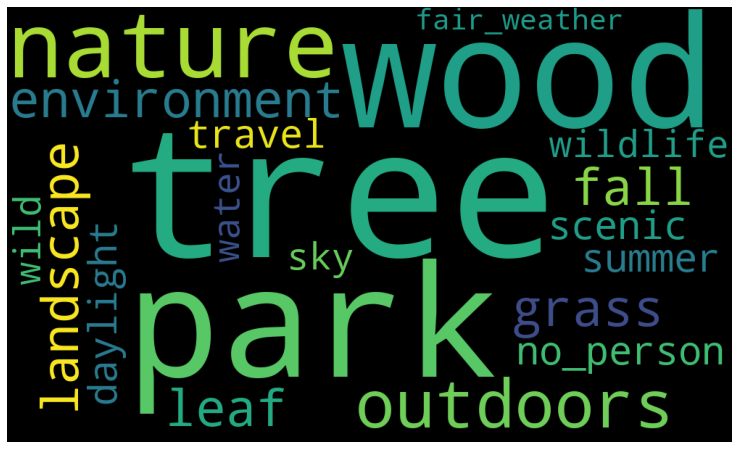}}
  \hfill
  \subfigure[Child]{\includegraphics{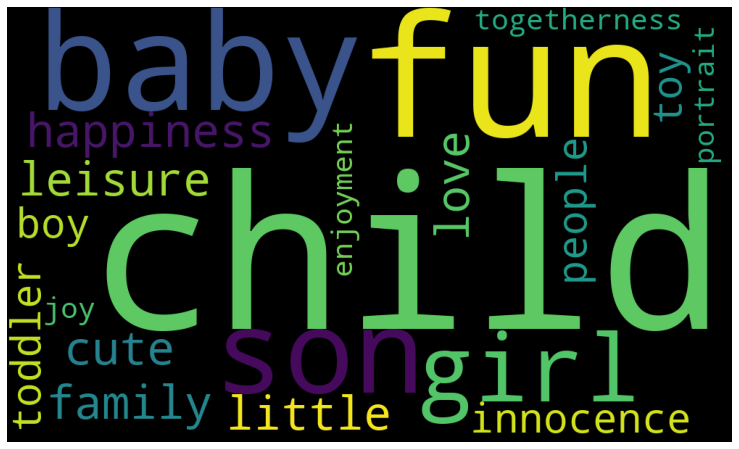}}
  \hfill
  \subfigure[Performance]{\includegraphics{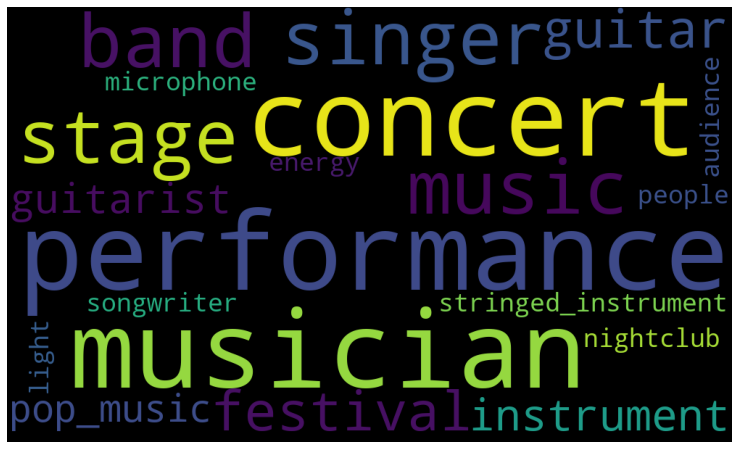}}
  \hfill
  \subfigure[Business]{\includegraphics{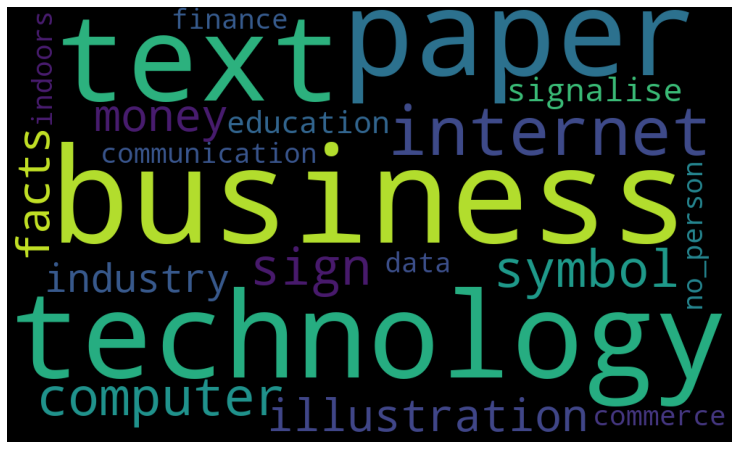}}
  \hfill
  \subfigure[Fashion]{\includegraphics{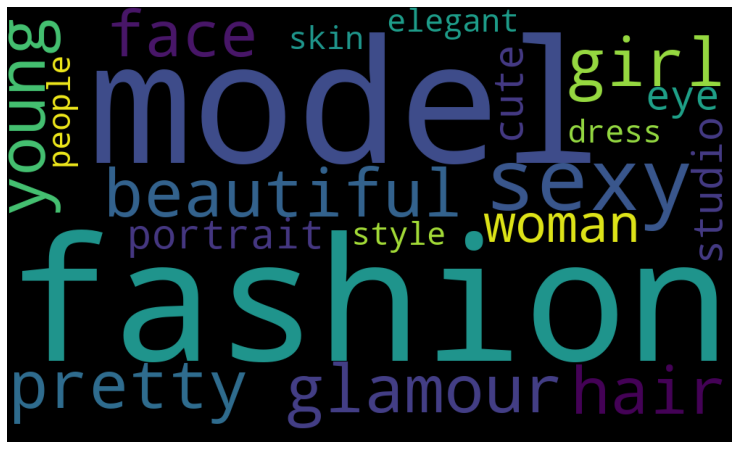}}
  \caption{Tag clouds for Topics Nature, Child, Performance, Business, and Fashion}
\label{fig:topics}
\end{figure*}

The topics extracted from NMF are nameless and thus not quickly understandable. Hence, we name the $20$ topics manually since the names will be used in the generated explanations.~Figure \ref{fig:topics} shows tag clouds of the top $20$ tags for five different topics.~The font size indicates relative significance, i.e.~the most descriptive tag is displayed as the largest.~For instance, the top five descriptive tags of the topic \textit{Nature} are $\left \{tree, park, wood, nature, outdoors\right \}$. 
%
\begin{figure}[htb]
  \centering
  {\includegraphics[scale=0.4]{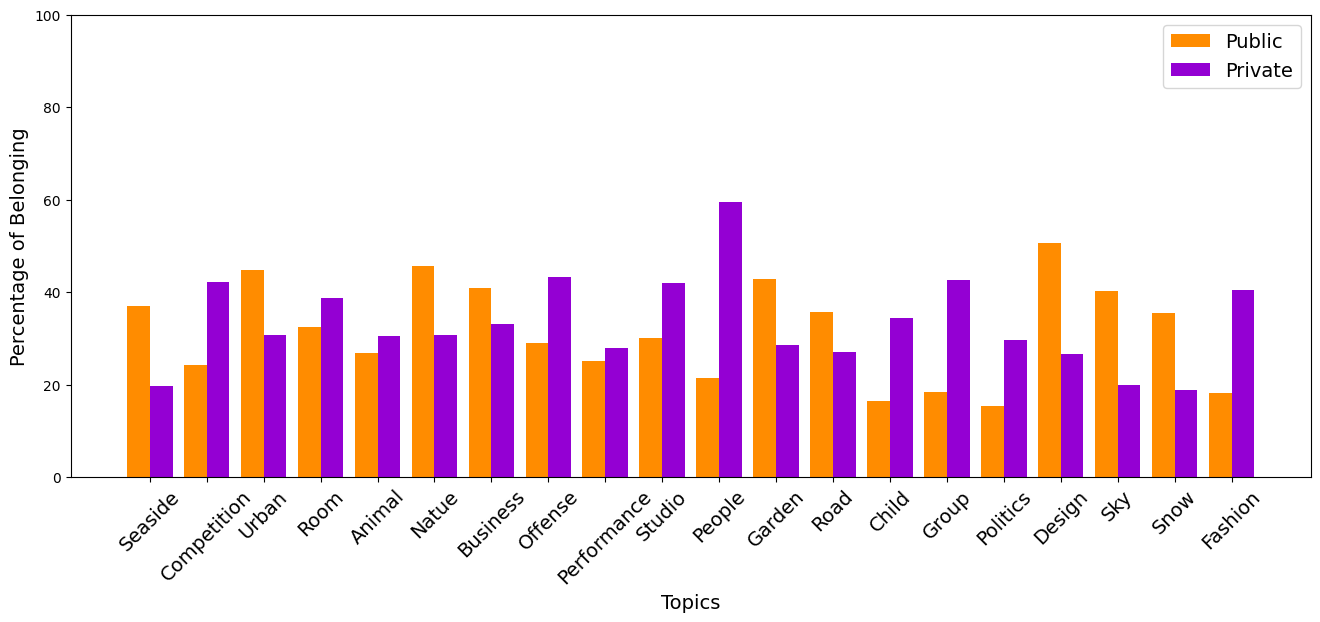}}
  \caption{Percentage of occurrence of each topic in private and public images}
  \label{fig:topic_class_freq}
\end{figure}
%

\section{Generating Explanations from Topics} 
\label{sec-generateexp}
We use the Random Forest algorithm to make image privacy predictions (Step $4$).~The question may arise whether certain topics might be more often associated with public or private images.~If so, that may make it relatively easy to explain the privacy predictions.~Figure \ref{fig:topic_class_freq} shows the share of each topic associated with private and public images.~Some topics such as \textit{People} are associated more frequently with the private class whereas others like \textit{Sky} are more often associated with the public class.~However, despite some topics being more frequently associated with a specific class, the topics ultimately do not have an explicit class to which they belong.~Therefore, a topic by itself does not directly signal a certain class, and as such it is not straightforward to generate an explanation for the decision by simply looking at a topic's class.

\subsection{Computing the Contribution of Topics using the TreeExplainer Model}
The TreeExplainer~\cite{lundberg2020local2global} is the implementation of the SHAP (SHapley Additive exPlanations) approach \cite{lundberg2017unified}, which can be used to understand how an ML model arrived at its prediction.~The TreeExplainer model provides the computation of local explanations based on Shapley values in polynomial time.~The model computes the contribution of each feature to a prediction, taking into account the interactions between features using tree-based models such as the Random Forest algorithm.~In this study, each feature corresponds to a topic.~Not all topics have an equal contribution to a class prediction: a topic can push the prediction higher (positive SHAP value) or lower (negative SHAP value), and their magnitude can differ.~An ML model concludes its prediction by taking into account the contribution of each topic.~This is useful in interpreting how the model works.~One way to create explanations would be to display all these values to the user.~However, as the number of topics increases, showing them all to the end user would be cumbersome and confusing.~Therefore, we start with the TreeExplainer output and then interpret it to make it understandable for the end user.

We are interested in identifying topics that are useful in explaining the content of the image at hand.~For example, for a given image, a large positive SHAP value might be assigned to a topic because the image is related to that topic.~But, it might also be the case that a large negative value is assigned to a topic that is unrelated to the image.~The latter shows the model made a decision based on the fact that the image did not exhibit the properties associated with this topic.~While useful to understand the model, this information is difficult to understand and may be unnecessary to show to the user.~Hence, we need to carefully decide how to use the SHAP values when creating the explanations. 

\subsection{Identifying Explanation Categories of Images}
Our privacy assistant generates human-understandable explanations by first reducing the number of topics in the TreeExplainer model output.~We bring structure to the explanations through the use of categories, thereby reducing complexity and making it easier to understand for users.~We make a distinction between explanations generated for end users and those for agents as they have different needs.~For end users, our explanation consists of two components, namely a short, explanatory text and a visual representation of the most important topics associated with the image.~Utilizing both a textual and a visual representation ensures the explanation is intuitive and easier to understand.~The image's category remains a key input though as each category has its own distinct explanatory text and visual representation.~In contrast, for agents, the sole explanation is the image category as no other information is necessary. 

First of all, following extensive manual work and analysis of the topics' contributions to the decisions, three distinct image categories were identified, namely \textit{Dominant}, \textit{Opposing}, and \textit{Collaborative}.~Images not definitively belonging to any of the three categories are grouped in a separate category called \textit{Weak}.~Second, our approach generates explanatory text for each explanation based on the explanation category.~That is, the explanatory text function has different patterns for different explanation categories in order to explain the visual representation and the relations between topics.~Third, our explanation approach provides a visual representation to indicate whether the image is public or private based on the topics and tags displayed.~Depending on the image's explanation category, it visualizes one or more topics with the related image tags in a circle labeled with the name of the topic(s).

The four explanation categories are further explained below and example explanations are given.

\mypara{Dominant:}An image belongs to the \textit{Dominant} category when the contribution of one topic is decisive for the class prediction.~In this case, a topic makes a relatively high contribution compared to all other topics of the image.~To determine whether a contribution is considered decisive for the prediction, a lower bound (\textit{db}) of $0.7$ is used. Hence, when there is a topic whose contribution exceeds this threshold, the image is considered to belong to the Dominant category.  

\begin{figure}[htb]
	\centering
	\includegraphics[scale=0.65]{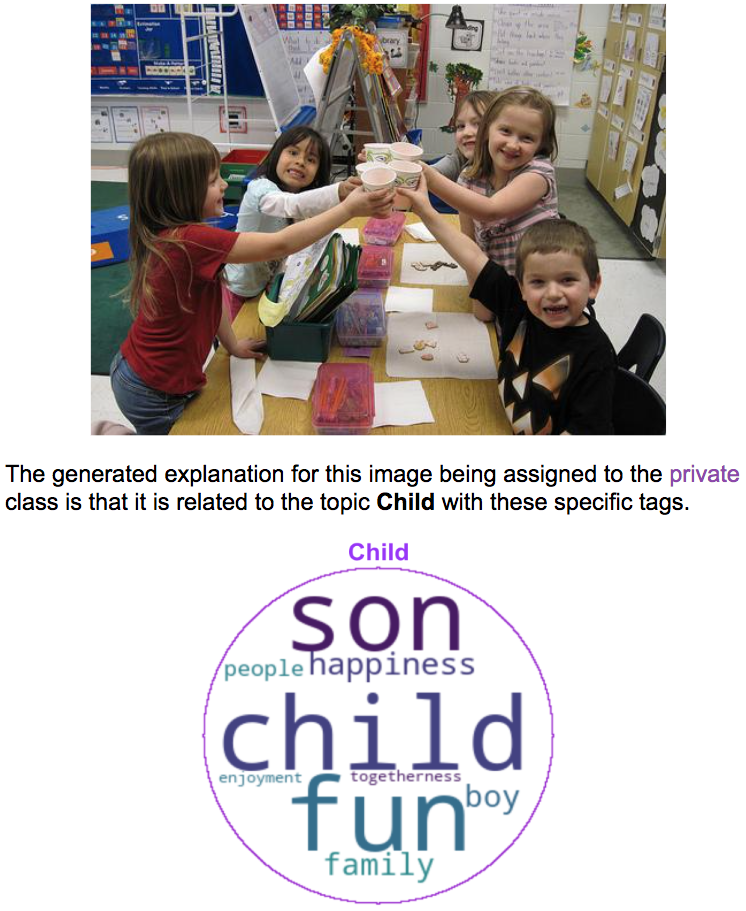}
	\caption{Example image annotated as private and its generated explanation with the topic Child (Dominant category)}
	\label{fig:cat_dominant}
\end{figure}

If the image is categorized as \textit{Dominant}, the generated explanation for a user contains only the dominant topic in both the text pattern and the visual representation.~The pattern of the explanatory text for the Dominant category is as follows: ``The generated explanation for the image being assigned to the public/private class is that it is related to the \textit{topic x} with the specific tags".~Figure \ref{fig:cat_dominant} shows an example of a \textit{Dominant} image identified as private by annotators.~For this example, the contribution of the topic Child is $8.6$ times larger than the topic with the next greatest contribution. 
%

\mypara{Opposing:}The topics associated with an image do not always conclusively point to whether the image should be classified as public or private, i.e.~opposing signals exist. Hence, an image belongs to the \textit{Opposing} category if it has topics whose contributions to a class prediction are in opposite directions and whose magnitudes are comparable \textit{and} sufficiently large. The latter is the case if the opposing contributions both exceed the lower bound (\textit{ob}) of $0.2$, i.e.~the contributions need to be sufficiently large. 
When an image has opposing topics whose contributions are in opposite directions, making a class prediction can be difficult. Hence, when this is the case, the explanation to the user should clearly indicate this and the pattern of the explanatory text for the Opposing category is therefore as follows: ``Even though it is related to the \textit{topic x} with the specific tags below (which signals the public/private class), it is also related to the \textit{topic y} and for that reason, it is classified as private/public".
%
\begin{figure}[htb]
	\centering
	\includegraphics[scale=0.65]{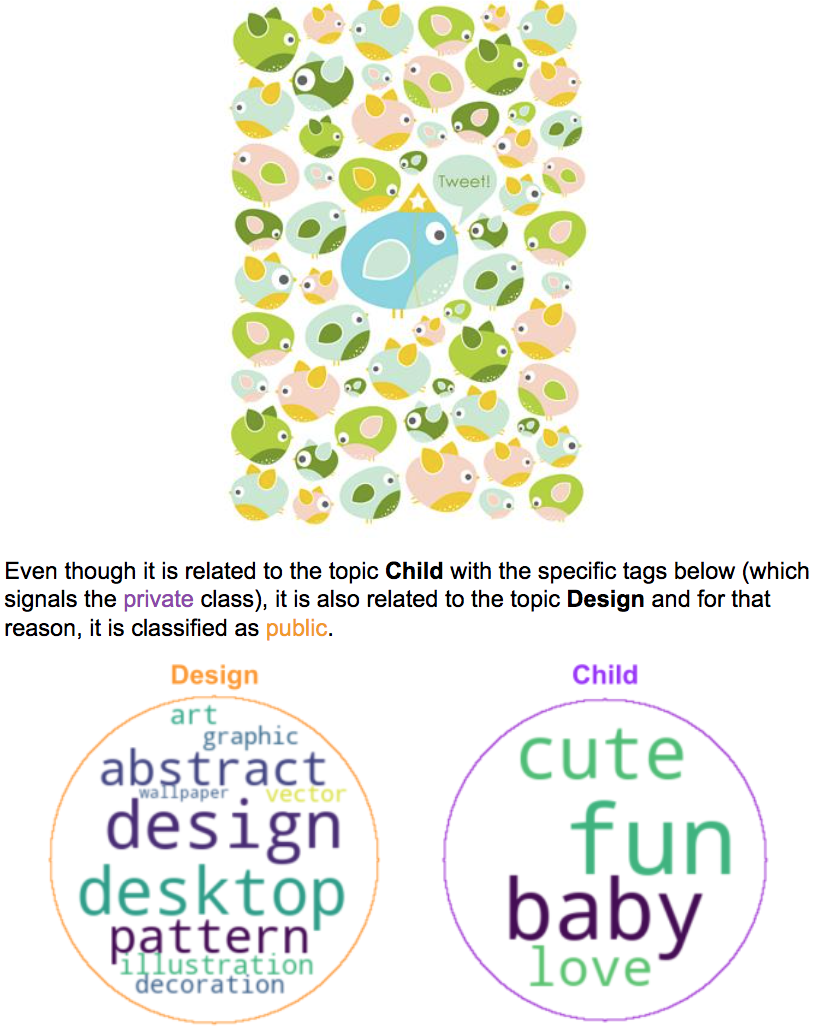}
	\caption{Example image annotated as public and its generated explanation with the topics Design and Child (Opposing category)}
	\label{fig:cat_Opposing}
\end{figure}
%
Figure \ref{fig:cat_Opposing} shows an example image identified as public by annotators. The generated explanation explicitly mentions the opposing topics, i.e.~Child and Design. The reason why the image has ultimately been classified as public is that Design's contribution higher outweighs Child's contribution pushing the decision lower. 

\mypara{Collaborative:}An image belongs to the \textit{Collaborative} category if \textit{enough}, though not \textit{all}, topics are \textit{decisively} pushing the decision in the same direction. Unlike the Dominant category, there is no single decisive topic. Instead, there are multiple topics whose combined collaborative contributions are decisive in classifying the image. If the sum of the collaborative contributions exceeds the Collaborative lower bound \textit{cb}, equal to $0.8$, the image belongs to the Collaborative category. Finally, the presence of a topic (or topics) pushing the decision in the opposite direction (of the collaborative direction) does not necessarily mean the image does not belong in the Collaborative category. As long as those opposing contributions are relatively minor, the image is still considered to be Collaborative rather than Opposing. The explanatory text of the Collaborative category includes the top three collaborative topics as follows: ``The generated explanation for the image being assigned to the public/private class is that it is related to the \textit{topics x, y,} and \textit{z} with these specific tags".
%
\begin{figure}[htb]
	\centering
	\includegraphics[scale=0.65]{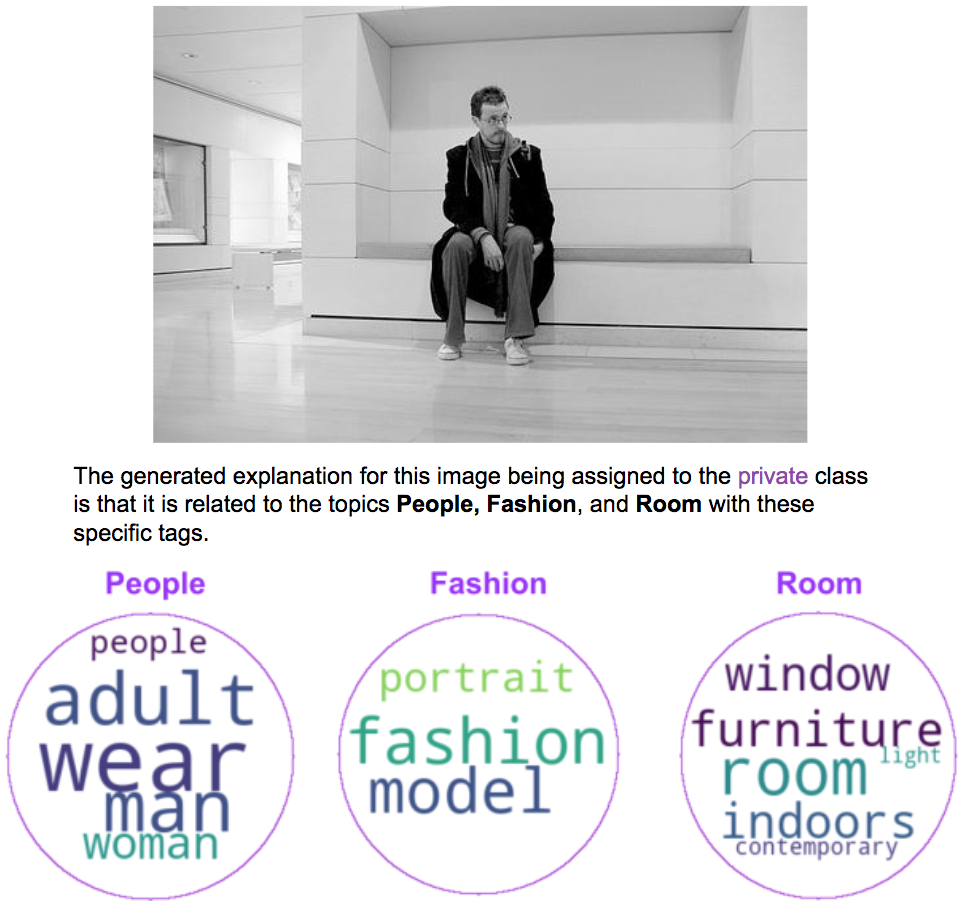}
	\caption{Example image annotated as private and its generated explanation with the topics People, Fashion, and Room (Collaborative category)}
	\label{fig:cat_collab}
\end{figure}
%
Figure \ref{fig:cat_collab} shows an example image identified as private. The generated explanation provides the topics People, Fashion, and Room (with relevant tags), which all push the decision to be private.

\begin{figure}[htb]
	\centering
	\includegraphics[scale=0.65]{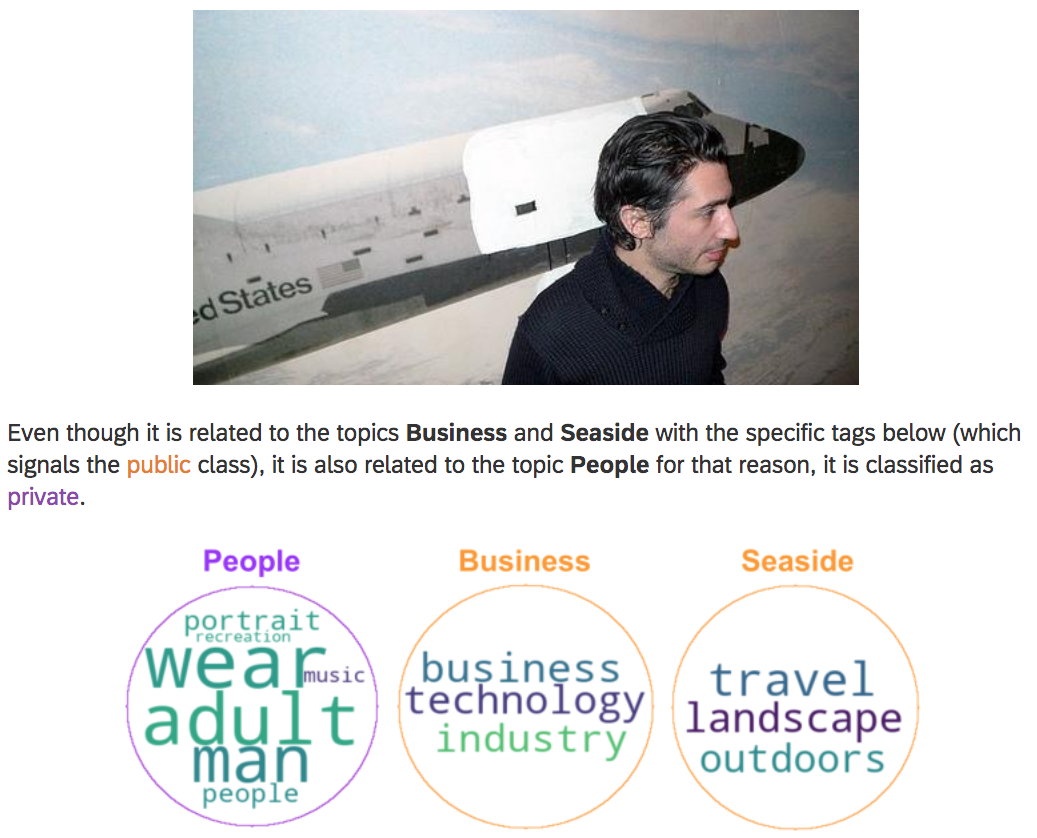}
	\caption{Example image annotated as private and its generated explanation with the topics People, Business, and Seaside (Weak category)}
	\label{fig:cat_Weak}
\end{figure}

\mypara{Weak:}Finally, it is also possible an image belongs to many topics with minor contributions.~Hence, it would not fall into any of the above three categories and its class can therefore not be explained as clearly as the others.~We call this category {\it Weak} and generate an explanation containing only the top contributing topics, i.e.~topics with relatively small contributions are not shown.~In doing so, our aim is to generate explanations with the most relevant and influential topics for the decision.~The explanatory text pattern for the Weak category, shown in Figure~\ref{fig:cat_Weak}, is the same as the explanatory text for the Opposing category.~Since none of the topics in this figure have a large contribution to the decision, no decisive decision can be made.  
%
%
\subsection{Computing Explanations}
Our method identifies four different explanation categories by utilizing the contributions of topics on the decision.~To evaluate the representativeness of the topics extracted using NMF with the images, we trained a Random Forest classifier where the images are represented as TF-IDF vectors of these topics.~We evaluate the performance (accuracy, precision, recall, and F1 score) on the Test set separately for private and public image classes.~This is critical as the consequences of misclassifying a private image could be more severe than misclassifying a public image.~Our findings demonstrate that the classifier achieves a precision of $87\%$ and $89\%$, recall of $90\%$ and $87\%$, and F1 score of $89\%$ and $88\%$ for private and public classes, respectively.~Overall, the classifier yields an accuracy of $88\%$ on the Test set, indicating the NMF-extracted topics are effective for privacy prediction.~It achieves state-of-the-art performance for image privacy prediction compared to existing approaches~\cite{tonge2020image,ayci2023uncertainty}. We implement our methodology using Python. The implementation details of this work are available at \url{https://github.com/aycignl/peak}.
\begin{algorithm}[]
\SetAlgoNoLine
  \caption{Generate Explanation for an Image}
  \label{alg:find_categories_topics}
  \SetKwInput{Input}{Input}
  \SetKwInOut{Output}{Output}
  
  \Input{$img$, a given image,\\
         $\qquad \quad \enspace T \in \mathbb{N}$, the number of topics,\\
         $\qquad \quad \enspace topic\_vector \in \mathbb{R}^{T}$, stores the SHAP values of the associated topics for \textit{img},\\
         $\qquad \quad \enspace norm\_vector \in \mathbb{R}^{T}$, stores the normalized values of the topic\_vector,\\
         $\qquad \quad \enspace sorted\_vector \in \mathbb{R}^{N}$, stores the indexes of the first $N$ topics sorted by $norm\_vector$ in descending order,\\
         $\qquad \quad \enspace \left \{db, ob, cb \right \} \in \left [0,1 \right ]$, lower bounds of Dominant, Opposing, and Collaborative categories, respectively, \\
         $\qquad \quad \enspace \text{TP} \text{ and } \text{TG}$, topics and tags spaces, respectively, \\
         $\qquad \quad \enspace \text{i2t}: \mathbb{Z}^{+} \rightarrow TG$, index to topic function,\\
         $\qquad \quad \enspace \text{et:} \sum^{*} \times 2^{TP} \times 2^{TP} \rightarrow \sum^{*}$, \text{explanatory text function},\\
         $\qquad \quad \enspace \text{im2t}: I \rightarrow 2^{TG}$, image to tag function,\\
         $\qquad \quad \enspace \text{tp2t}: TP \rightarrow 2^{TG}$, topic to tag function.\\
         }
    \Output{$\text{explanation:} \sum^{*} \times \sum^{*} \times 2^{TP\times TG}$}  
    \BlankLine
    \nl $\text{explanation} \leftarrow \left (\emptyset, \emptyset, \emptyset\right )$\\
        $c\_sum^{+}, c\_sum^{-} \leftarrow 0, 0$\\
        $c\_topics^{+}, c\_topics^{-} \leftarrow \left [  \right ], \left [  \right ]$\\
        $o\_topics^{+}, o\_topics^{-} \leftarrow \left [  \right ], \left [  \right ]$\\

    \nl  \If{$norm\_vector\left [ sorted\_vector\left [1 \right ]\right] \geq db $}{
    \nl     $d\_topic \leftarrow i2t(sorted\_vector\left [1 \right ])$\\
    \nl     $d\_tags \leftarrow im2t(img) \cap tp2t(d\_topic)$
    \nl     $d\_topic\_tags \leftarrow (d\_topic, d\_tags)$\\
    \nl     $ \text{explanation} \leftarrow  (``dominant", et(``dominant", \left \{ d\_topic  \right \}, \emptyset), d\_topic\_tags) $
    }
    \nl   \Else{
    \nl      \For{n = 1 \textbf{to} {N}}{%
    \nl     \If{$topic\_vector[sorted\_vector\left [n \right ]] > 0$}{
    \nl       $c\_sum^{+} \leftarrow c\_sum^{+} + norm\_vector[sorted\_vector\left [n \right ]]$\\
    \nl       $c\_topics^{+} \leftarrow c\_topics^{+} \cup \left \{ i2t(sorted\_vector\left [n \right ] )\right \}$\\
    \nl       \If{$norm\_vector[sorted\_vector\left [n \right ]] \geq ob$}{
    \nl         $o\_topics^{+} \leftarrow o\_topics^{+} \cup \left \{ i2t(sorted\_vector\left [n \right ]) \right \}$}}
          }
        }
\end{algorithm}

\begin{algorithm}[]
\setcounter{AlgoLine}{13}
\SetAlgoNoLine
    \Indp \Indp \nl  \ElseIf{$topic\_vector[sorted\_vector\left [n \right ]] < 0$}{
    \nl       $c\_sum^{-} \leftarrow c\_sum^{-} + norm\_vector[sorted\_vector\left [n \right ]]$\\
    \nl       $c\_topics^{-} \leftarrow c\_topics^{-} \cup \left \{ i2t(sorted\_vector\left [n \right ]) \right \}$\\    
    \nl       \If{$norm\_vector[sorted\_vector\left [n \right ]] \geq ob$}{
    \nl         $o\_topics^{-} \leftarrow o\_topics^{-} \cup \left \{ i2t(sorted\_vector\left [n \right ]) \right \}$
                }  
            } \Indm
    \nl  \If{$\left | o\_topics^{+} \right | \in \mathbb{Z}^{+} \text{ and } \left | o\_topics^{-} \right | \in \mathbb{Z}^{+}$}{
    \nl     $o\_topics\_tags^{+} \leftarrow \bigcup_{i=1}^{\left | o\_topics^{+} \right |} \left \{ \left ( o\_topics^{+}[i], im2t(img) \cap tp2t(cf\_topic^{+}[i]) \right )\right \}$\\
    \nl     $o\_topics\_tags^{-} \leftarrow \bigcup_{i=1}^{\left | o\_topics^{-} \right |} \left \{ \left ( o\_topics^{-}[i], im2t(img) \cap tp2t(cf\_topic^{-}[i]) \right )\right \}$\\
    \nl     $explanation \leftarrow  (``opposing", et(``opposing", o\_topics^{+}, o\_topics^{-}), \newline o\_topics\_tags^{+} \cup o\_topics\_tags^{-}) $\\
    }
    \nl  \ElseIf{$c\_sum^{+} \geq cb$}{
    \nl    $c\_topics\_tags \leftarrow \bigcup_{i=1}^{3} \left \{ \left ( c\_topics^{+}[i], im2t(img) \cap tp2t(cl\_topic^{+}[i]) \right )\right \}$\\
    \nl    $explanation \leftarrow  (``collaborative", et(``collaborative", c\_topics^{+}, \emptyset), c\_topics\_tags) $}
    \nl  \ElseIf{$c\_sum^{-} \geq cb$}{
    \nl   $c\_topics\_tags \leftarrow \bigcup_{i=1}^{3} \left \{ \left ( c\_topics^{-}[i], im2t(img) \cap tp2t(c\_topic^{-}[i]) \right )\right \}$\\
    \nl    $explanation \leftarrow  (``collaborative", et(``collaborative", \emptyset, c\_topics^{-}), c\_topics\_tags) $}
    \nl   \Else{
    \nl    $w\_topics \leftarrow i2t(sorted\_vector\left [:3 \right ])$\\
    \nl    $w\_topics\_tags \leftarrow \bigcup_{i=1}^{3} \left \{ \left ( w\_topics[i], im2t(img) \cap tp2t(w\_topics[i]) \right ) \right \}$\\
    \nl    $explanation \leftarrow  (``weak", et(``weak", w\_topics, \emptyset), w\_topics\_tags)$}
\end{algorithm}

Algorithm \ref{alg:find_categories_topics} shows how our proposed approach categorizes an image, gives appropriate topics and tags to be used in the user explanation, and ultimately generates an explanation (specific to each category) why the image is classified as public or private.~In this algorithm, $T$ is the number of topics associated with an image.~$topic\_vector$ is a vector containing the SHAP values of the associated topics for the image.~Additionally, we normalize these values by dividing each value by the sum of the absolute values of all image topics and storing the normalized values of $topic\_vector$ in $norm\_vector$.~In a final step, the algorithm creates a sorted vector ($sorted\_vector$) containing the first $N$ topics of the $norm\_vector$, sorted in descending order.

\textit{db, ob,} and \textit{cb} are the respective lower bounds, as previously mentioned in the category definitions, with regard to deciding whether a given image belongs to the Dominant, Opposing, or Collaborative category.~A number of functions are needed in order to generate category-specific output explanations, with $TP$ and $TG$ as respectively the topics and tags spaces, i.e.~all explored $N$ topics and associated tags.~The function $i2t$ returns topic names from the indexes of $sorted\_vector$, e.g. the first element is topic Child.~The function $et$ is the explanatory text function that creates the text pattern used in the explanation.~It uses as input the image's category and its explanation topic(s), which are the most relevant topics for the prediction.~For instance, only the Dominant topic for Dominant images.~The $im2t$ function returns the tags associated with a given image whilst the $tp2t$ function returns the tags associated with a given topic.~The functions together are needed to ultimately find the common tags associated with \textit{both} the image and a certain topic.~Finally, Algorithm \ref{alg:find_categories_topics} features as output $explanation$, which consists of three components, namely the image's category name, the generated explanatory text from the $et$ function and the explanation topic(s) with associated tags.

Several variables are initialized (line $1$), starting with $explanation$. Additionally, in order to be able to determine whether an image belongs to the Collaborative category, the algorithm initializes two variables, namely the total sum of the contributions of topics pushing the prediction higher (i.e.~$c\_sum^{+}$) or lower (i.e.~$c\_sum^{-}$). Secondly, two variables used to store the explanation topics for Collaborative images are initialized, respectively $c\_topics^{+}$ and $c\_topics^{-}$ for higher- and lower pushing topics. Finally, the algorithm initializes two variables containing lists of topics that push the prediction higher (i.e.~$o\_topics^{+}$) or lower (i.e.~$o\_topics^{-}$). These variables are used to assess whether a given image belongs to the Opposing category. 

Algorithm \ref{alg:find_categories_topics} first checks whether the image is in the Dominant category.~When there is a single topic whose contribution is disproportionately high compared to the other topics, the image is deemed to be Dominant.~Hence, if the normalized value of the first topic in the $sorted\_vector$ (i.e.~the topic with the largest contribution) is equal to or greater than the Dominant's bound, the image is considered Dominant (line $2$).~Then the $i2t$ function subsequently returns the Dominant topic name ($d\_topic$, line $3$) whilst its associated tags $d\_tags$ are returned from the intersection of the $im2t$ and $t2t$ functions (line $4$).~The two are then stored in $d\_topic\_tags$ (line $5$).~Finally, the algorithm outputs the explanation using the generated inputs (line $6$).

If the image is not Dominant (line $7$), variables for the Collaborative and Opposing categories are constructed by cycling through each topic up to $N$ (line $8$) and based on the direction of the topic's contribution to the prediction. If a topic is pushing the decision higher (line $9$), then its normalized value (i.e.~its contribution) is added to $c\_sum^{+}$ (line $10$) and its name is added to $c\_topics^{+}$ (line $12$). Additionally, if the topic's normalized value is equal to or greater than the Opposing's bound, the topic is appended to the $o\_topics^{+}$ set (lines $12-13$). In contrast, if a topic is instead pushing the decision lower (line $14$), its value is added to $c\_sum^{-}$ (line $15$) and its name is appended to the $c\_topics^{-}$ set (line $16$). Moreover, if the topic's normalized value is equal to or greater than the Opposing's bound (line $17$), the topic is appended to the $o\_topics^{-}$ list (line $18$). Once all topics of a given image have been cycled through and the variables have been assigned values, the algorithm is able to assign the image to a category (i.e.~Collaborative or Opposing) based on its features.

If the image has at least one topic pushing the decision \textit{sufficiently} higher (i.e.~$o\_topics^{+}$ contains a topic) \textit{and} at least one topic pushing the decision \textit{sufficiently} lower (i.e.~$o\_topics^{-}$ contains a topic), the image is categorized as Opposing (line $19$) and the algorithm outputs the appropriate explanation using the most important topics and topic tags (lines $20$-$22$). 

If the image is not in the Opposing category, the algorithm instead checks if its topics are acting in a collaborative manner. It first checks whether the sum of all contributions pushing the decision higher ($c\_sum^{+}$) is equal to or greater than the Collaborative's bound (line $23$). If true, it means there are \textit{enough} topics pushing the decision \textit{decisively} higher. The top three most contributing topics ($c\_topics^{+}$) and associated tags are subsequently returned (lines $24-25$) and the explanation is outputted. On the other hand, if the image's topics are not collaboratively pushing the decision higher, the algorithm checks whether the opposite is true. Hence, if topics are together pushing the decision \textit{decisively} lower and $c\_sum^{-}$ is thus equal to or greater than the Collaborative's bound, the image is considered Collaborative and the appropriate explanation is generated (lines $26-28$). 

Finally, if the image does not belong to any of these categories, the algorithm considers the image as Weak and an explanation featuring the most relevant three topics and associated tags is generated (lines $29-32$). The algorithm ultimately generates as output an explanation of why the image has been classified as public or private. 

\section{Evaluation}
We evaluate the performance of our proposed system in terms of its contribution to preserving privacy using explanations. 

\subsection{User Study}\label{sec-userstudy}
We first aim to answer the following research question:
\begin{description} 
    \item[RQ1] Are the generated explanations by \peak~sufficient, satisfying, and understandable for humans?
\end{description}
We perform an online user study to evaluate our proposed explanation model in terms of sufficiency, satisfaction, and understanding. We conduct a pilot study (with $n=5$ participants) before the real study to test whether the study is understandable. Based on the comments during the pilot, we improved the initial description of the study and reworded one question.

Our user study has three phases. In the first phase, we present a plain language statement that describes the study and a consent form. The second phase is meant to explain the study over an example, wherein we show an image, its generated explanation, and the three questions that will be asked to the participant. Finally, in the third phase, each participant is exposed to $16$ images with generated explanations in random order. Two of these images deliberately provide irrelevant explanations so that we can differentiate the participants that are attentive during the survey. Thus, these questions are meant to filter out the participants who are not focused.  Such users are removed from the analysis.  

In order to examine our explanation model, we personalize the \textit{Explanation Satisfaction Scale} proposed by Hoffman {\it et al.} \cite{hoffman2018metrics}. We ask participants to rank the following questions:

\begin{enumerate}
    \item This explanation that the algorithm produces has SUFFICIENT DETAIL.
    \item This explanation produced by the algorithm is SATISFYING.
    \item From this explanation, I UNDERSTAND why an image has been identified as private or public.
\end{enumerate}

Each factor is accompanied by a $5-$point Likert scale (Strongly agree $= 5$, Somewhat agree $= 4$, Neither agree nor disagree $= 3$, Somewhat disagree $= 2$, Strongly disagree $= 1$). In the final phase, participants responded to anonymously collected demographic questions (age, gender, and education level) and optionally provided free-form text for comments/feedback. We designed our user study using the Qualtrics online survey tool \footnote{https://www.qualtrics.com}.  

\subsubsection{Participants}
A total of $57$ participants responded to questions but we excluded $12$ of them who did not catch the check questions properly. $64\%$ of the remaining $45$ participants were male and $36\%$ were female. $26$ participants were between 25-34 years old, $14$ were between 18-24, $4$ were between 35-44, and $1$ was between 55-64. In terms of the highest degree of education, $19$ of them had a Master’s degree, $11$ of them had Bachelor’s degree, $6$ of them were High school graduates, $5$ of them attended Some college (1-4 years, no degree), $2$ of them had Doctorate degree, and $2$ of them had Professional school degree (MD, DDC, JD, etc). This demographic is well-balanced on gender, age group, and education.

\subsubsection{Results}
\begin{figure*}[htb]
  \centering
  \SetFigLayout{1}{1}
    {\includegraphics{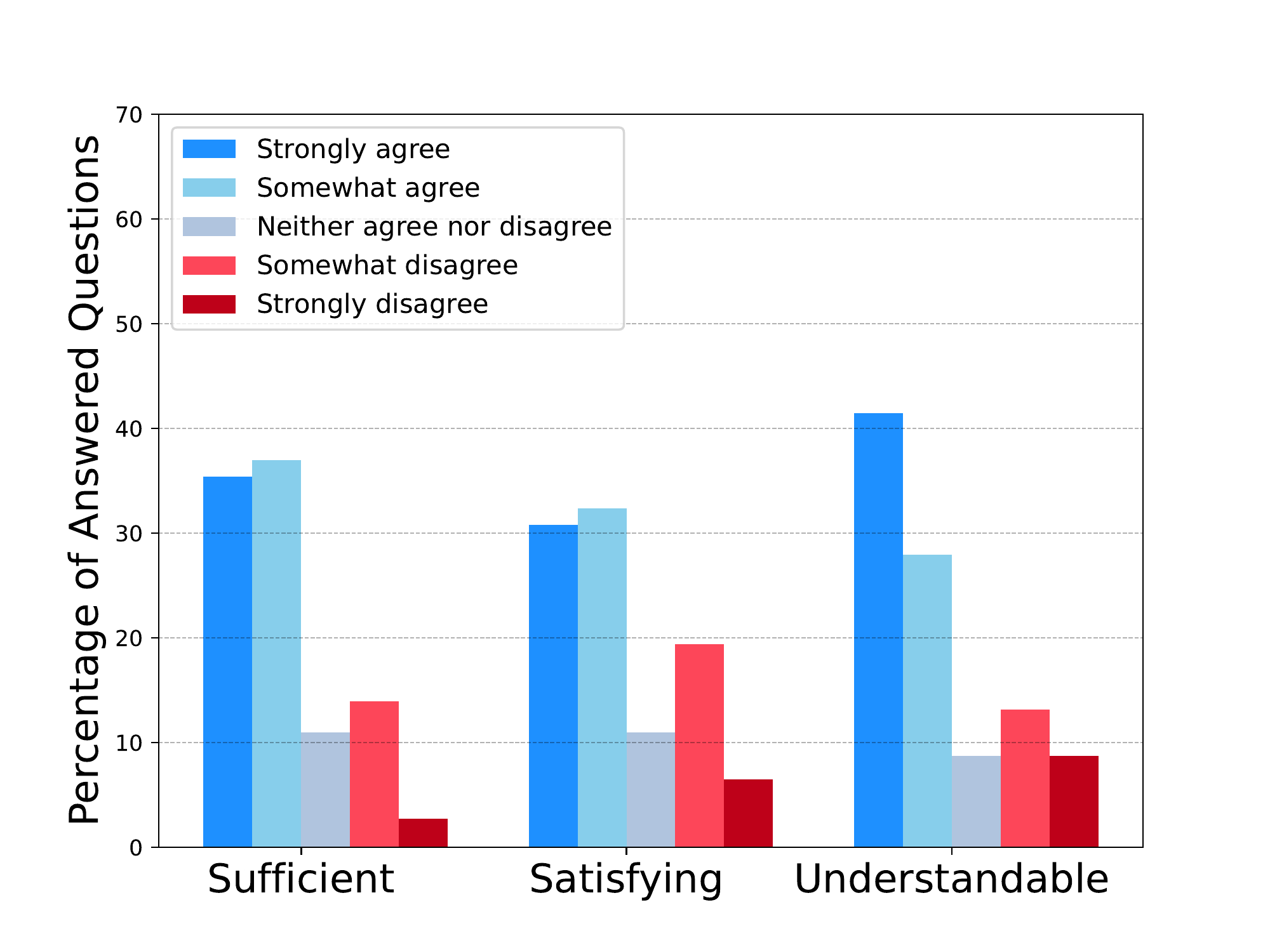}}
  \caption{The answers for all questions} 
\label{fig:overall_answers}
\end{figure*}

\begin{figure*}[htb]
  \centering
  \SetFigLayout{1}{2}
  \subfigure[Public]{\includegraphics{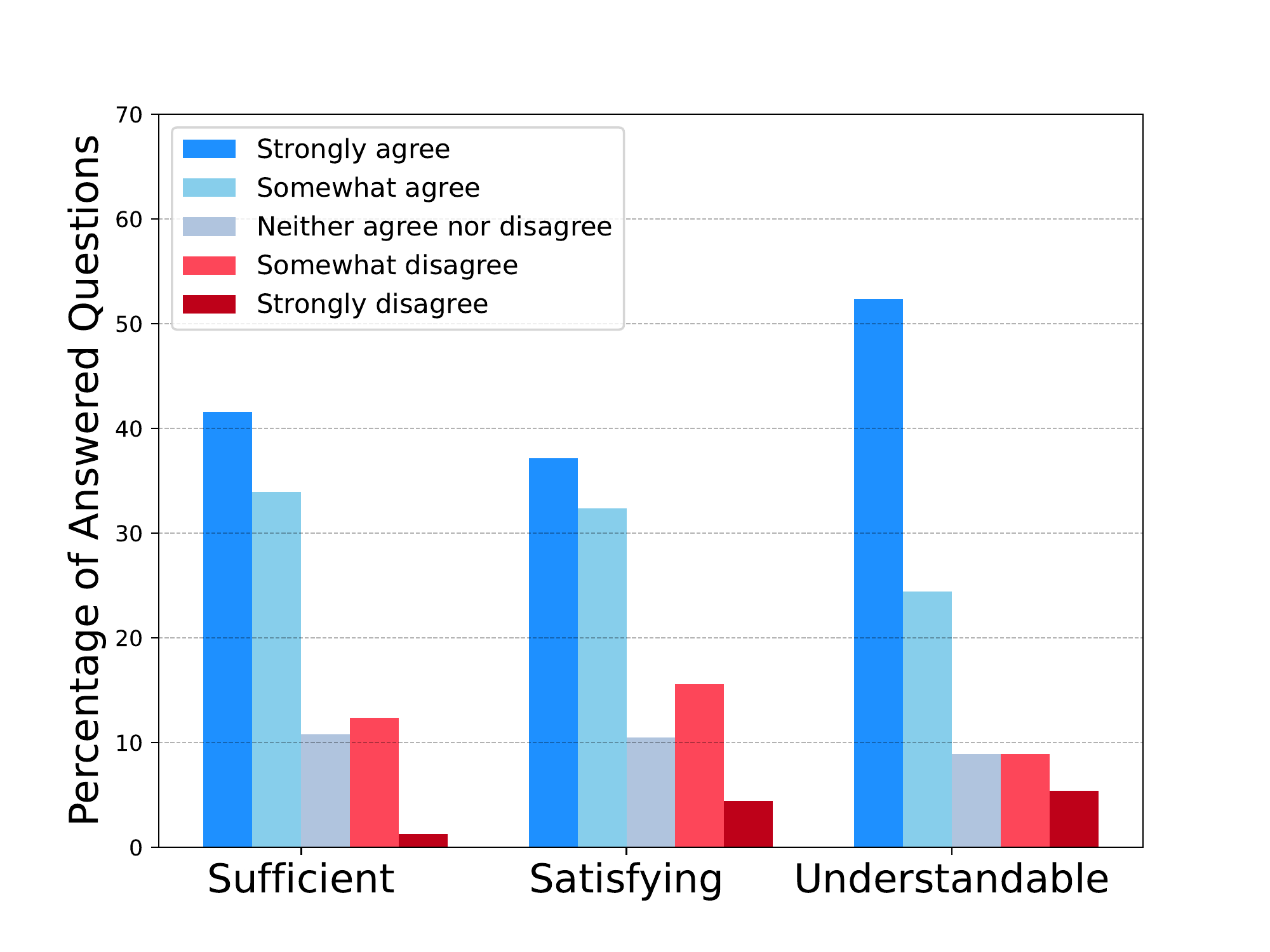}\label{fig:us_classes_public}}
  \hfill
  \subfigure[Private]{\includegraphics{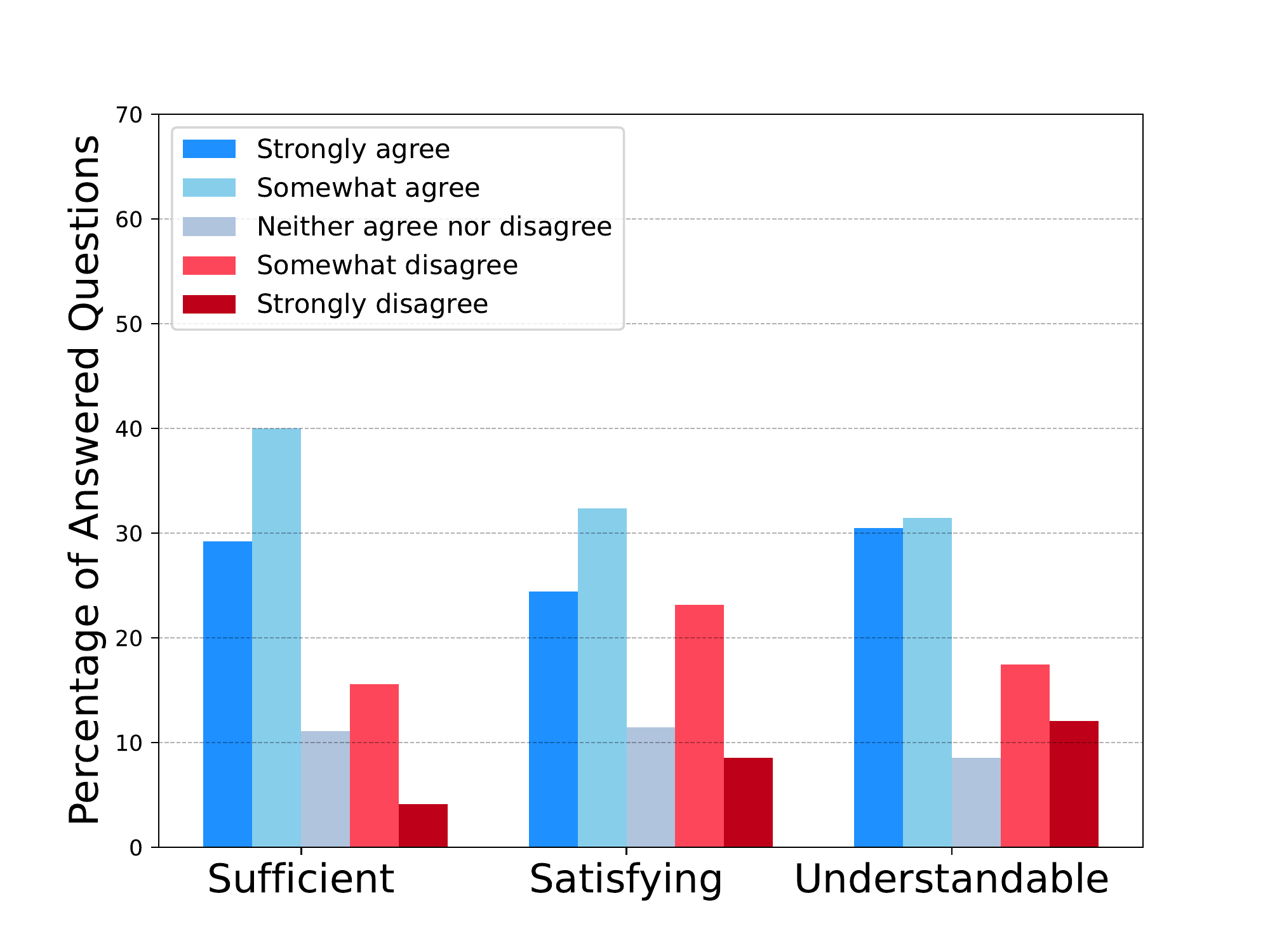} \label{fig:us_classes_private}}
  \caption{The answers for the questions with respect to the classes} 
  \label{fig:us_classes}
\end{figure*}

\begin{figure}[h]
  \centering
  \SetFigLayout{2}{2}
  \subfigure[Dominant]{\includegraphics{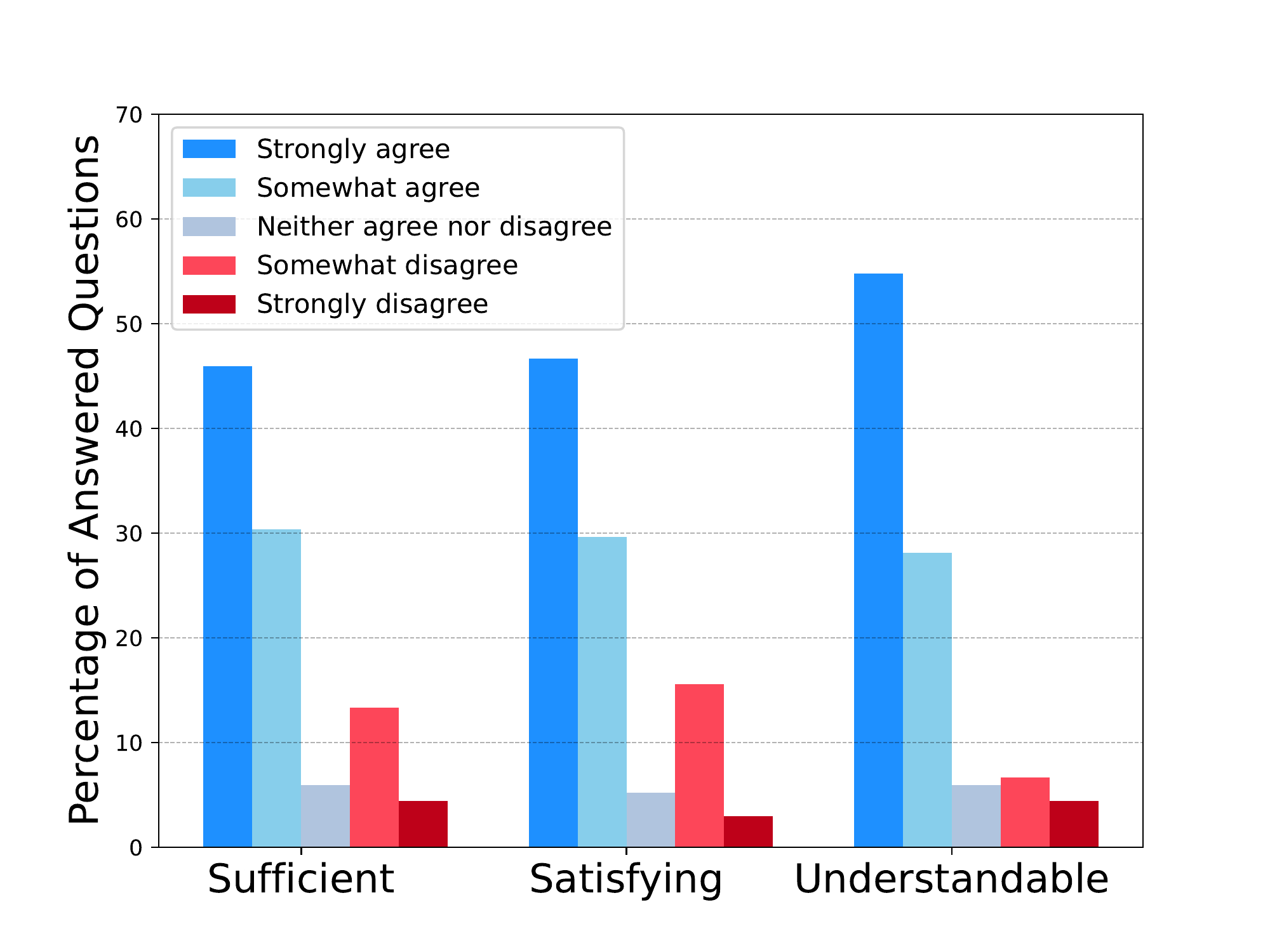}\label{fig:us_categories_dominant}}
  \hfill
  \subfigure[Opposing]{\includegraphics{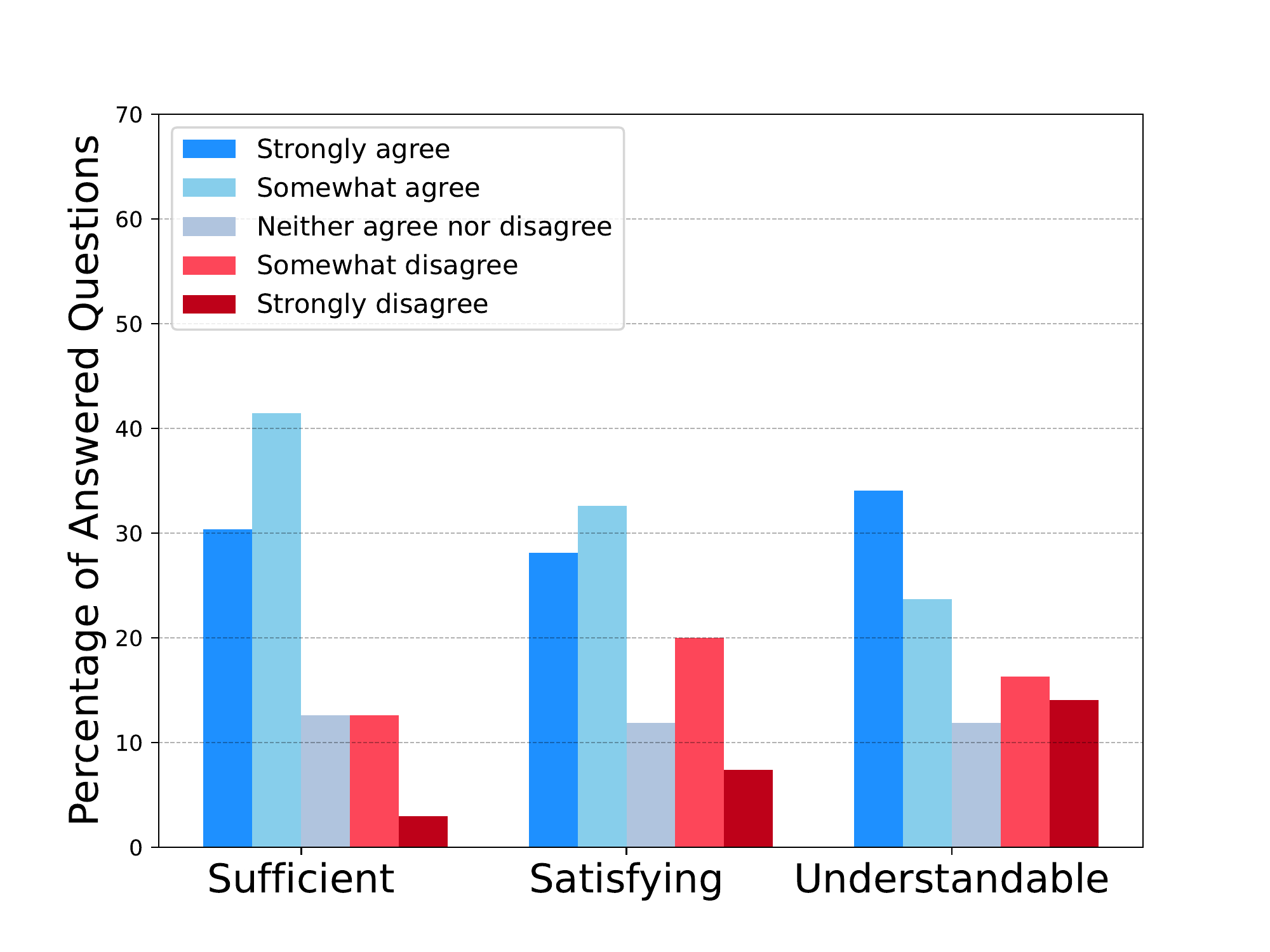}\label{fig:us_categories_Opposing}}
  \hfill
   \subfigure[Collaborative]{\includegraphics{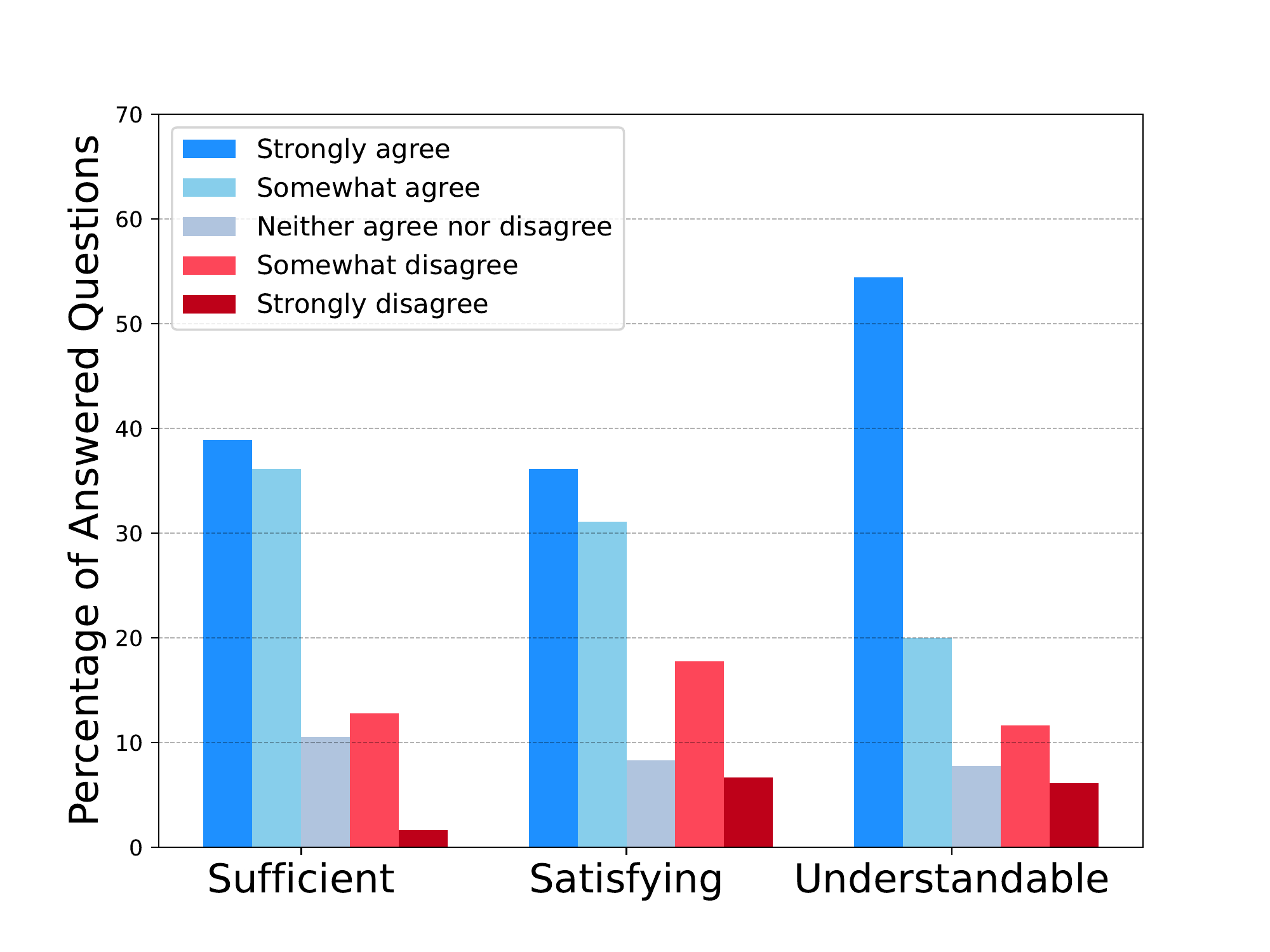}\label{fig:us_categories_collaborative}}
  \hfill
  \subfigure[Weak]{\includegraphics{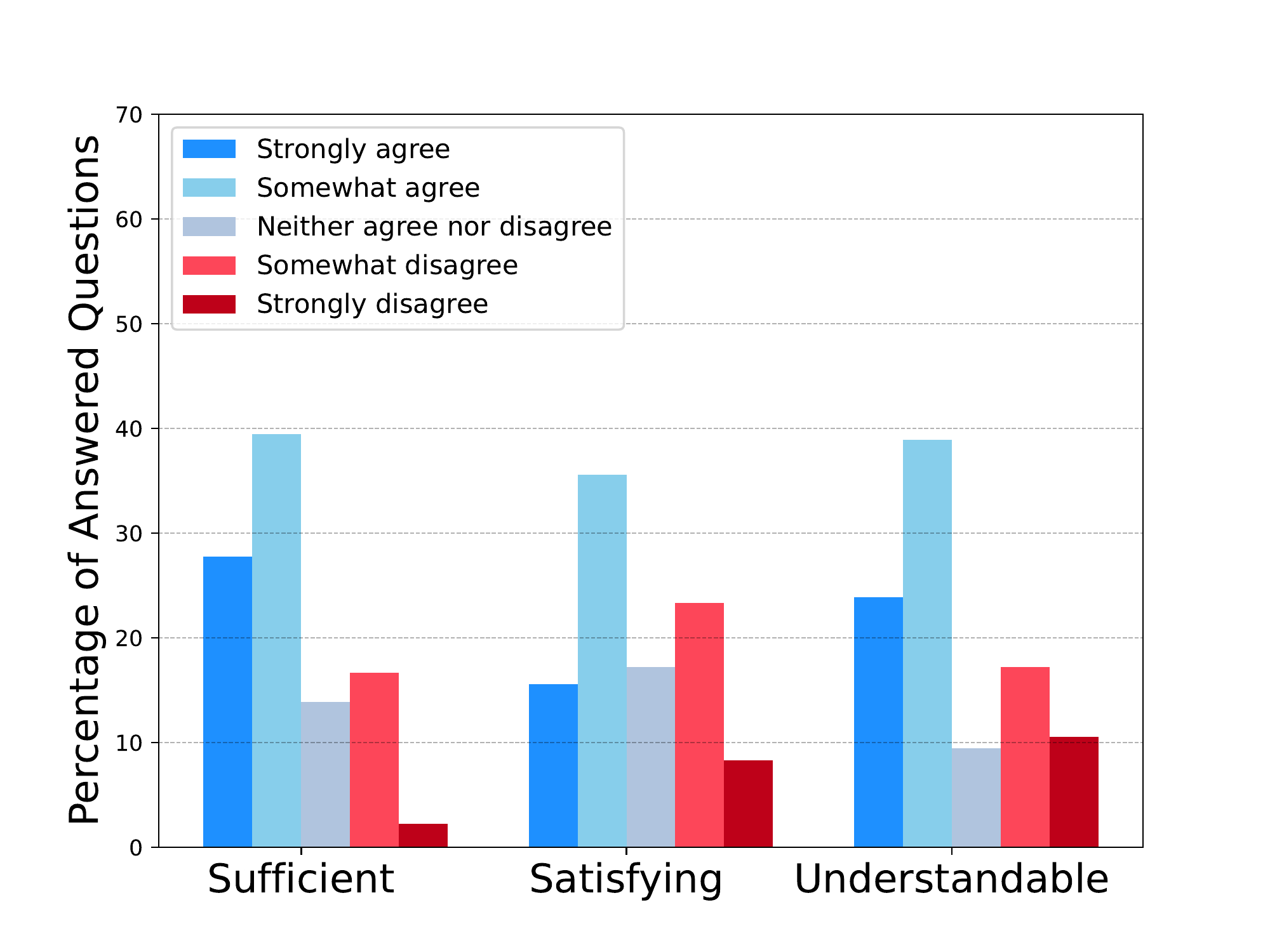}\label{fig:us_categories_Weak}}
  \caption{The answers for the questions with respect to the categories}
  \label{fig:us_categories}
\end{figure}

We show our results of the user study in Figures \ref{fig:overall_answers}, \ref{fig:us_classes} and \ref{fig:us_categories}, starting with the results for all images, then looking at the public and private classes specifically, and finally discussing the results for different explanation categories.

The results for all images (Figure \ref{fig:overall_answers}) are positive and show that the generated explanations are useful for and make sense to humans. Participants generally agreed the explanations were sufficiently detailed, they found the explanations satisfactory, and they understood why the images were labeled as private or public.

Firstly, explanations were deemed to be sufficiently detailed ($M = 3.88, SD = 1.12$) as over $70\%$ of respondents answered they either ``strongly agree" or ``somewhat agree", with the result statistically significant $\left ( \text{p-value} < 0.05 \right )$. Moreover, only a low proportion of respondents indicated they ``strongly disagree". 

Secondly, explanations were generally seen as satisfactory ($M = 3.62, SD = 1.28$) by respondents, a statistically significant result at the $0.05$ significance level. More than $60\%$ of participants answered in agreement and whilst there was some disagreement reported, this was largely only ``somewhat disagree".

Finally, the explanations for why images were labeled as public or private were generally deemed to be understandable by participants ($M = 3.8, SD = 1.33$), again statistically significant at the $0.05$ level. Out of all three dimensions, ``strongly agree" was the most commonly given answer at over $40\%$ of respondents, followed by close to $30\%$ of ``somewhat agree" answers. The range of answers was more varied however, as indicated by the larger standard deviation and a higher proportion of ``strongly disagree" compared to \textit{Sufficient} and \textit{Satisfying}.
%

Figure \ref{fig:us_classes} shows the distributions of answers for the survey questions with respect to the public and private classes. Figure \ref{fig:us_classes_public} indicates that participants found the explanations for public images to be more sufficient, more satisfying, and more understandable compared to the images labeled as private. For all three dimensions, participants agreed (``strongly agree" and ``somewhat agree" combined) more frequently for public images as opposed to private images. Additionally, ``strongly agree" was the most frequent answer for public images in contrast to private images where ``somewhat agree" is the most commonly given answer. For private images, there was more disagreement reported, with higher shares of ``somewhat disagree" and ``strongly disagree". For sufficiently detailed and satisfying, the differences between public and private images are not that large but for understandable there is quite a stark difference. For public images, more than half of respondents strongly agreed the explanations were understandable whereas for private images the share is only around $30\%$ and nearly equal to the number of ``somewhat agree" answers. Moreover, more respondents disagreed, and quite strongly so, for private images. A possible explanation for why participants generally answered ``strongly agree" more often for public images across all three dimensions may be because privacy and what are considered private images is subjective and more difficult to explain. Hence, respondents agree less often and their responses vary more for private images compared to public images.
%

Figure \ref{fig:us_categories} shows the distributions of answers to assess sufficiency, satisfaction, and understandability with respect to different categories (i.e., Dominant, Opposing, Collaborative, and Weak). Figure \ref{fig:us_categories_dominant} and \ref{fig:us_categories_collaborative} demonstrate when an explanation has a decisive topic or is composed of like-minded topics in the decision, participants agree that the explanations of images belonging to such categories are sufficiently detailed and satisfying. Additionally, participants understand why an image is identified as belonging to a certain class (private or public). On the other hand, compared to the Dominant and Collaborative categories, Figure \ref{fig:us_categories_Opposing} shows that participants score understanding ($M = 3.47, SD = 1.45$) of the decision less when an explanation has topics that have opposing forces in the decision. The images in this category have Opposing topics in terms of the contribution to the decisions. Thus, making a decision is not straightforward for the images belonging to the Opposing explanation category when compared to the Dominant and Collaborative categories. Moreover, Figure \ref{fig:us_categories_Weak} shows the results for the explanations of the images belonging to the Weak category. Even if participants are only moderately confident ($M = 3.27, SD = 1.22$) that the explanations are satisfying, they agree on the sufficiency of the explanations and understandability of a class decision based on the explanations. In light of these results from the user study, we answer RQ1 as participants find generated explanations by \peak~sufficient, satisfying, and understandable.

\subsection{Enhancing Privacy Assistants}
\label{sec:result_performance}
We also evaluate the performance of \peak~in terms of enhancing privacy assistants.~We aim to answer the following research question:
\begin{description} 
    \item[RQ2] Can the generated explanations of \peak~be used by personal assistants to improve their decision making?
\end{description}

\peak~can help personal assistants to classify images that are difficult to make a prediction about, e.g. low confidence predictions. It does so by dividing images into explanation categories, which allows it to uncover distinct and hidden relationships between topics for each category. The explanation categories are subsequently used to generate explanations for why an image is considered public or private. 

\peak~is used in conjunction with personal assistant \pure. \pure~is a personalized uncertainty-aware privacy assistant that helps its user to make privacy decisions \cite{ayci2023uncertainty}.~While \pure~makes a prediction (i.e.~\textit{public} or \textit{private}) for each image, it also captures the ambiguity of privacy by calculating a level of uncertainty for that prediction.~\pure~has access to a collection of data that has been labeled by various annotators (i.e.~PicAlert dataset).~Using only the visual characteristics of images, \pure~learns users' privacy preferences and creates a model.~To make a privacy decision, \pure~compares the uncertainty level to a threshold level provided by the user, and if the threshold is exceeded, \pure~delegates the privacy decision to the user.~Otherwise, \pure~uses its own prediction results (i.e.~share or not share). \pure~is a useful approach because uncertain images are not decided by the personal assistant, thus fewer mistakes are done.~However, this means the user has to decide on the uncertain images. \peak~allows privacy assistants to classify images accurately, ultimately resulting in fewer images having to be delegated to the user.~By doing so, the user can be more efficient while making decisions for fewer images.~Hence, \peak~decreases the cognitive load of the user.~Ultimately, the decision-making of the privacy assistants will have been improved without losing their performance.
%
\begin{figure}[]
    \centering
    \includegraphics[scale=0.55]{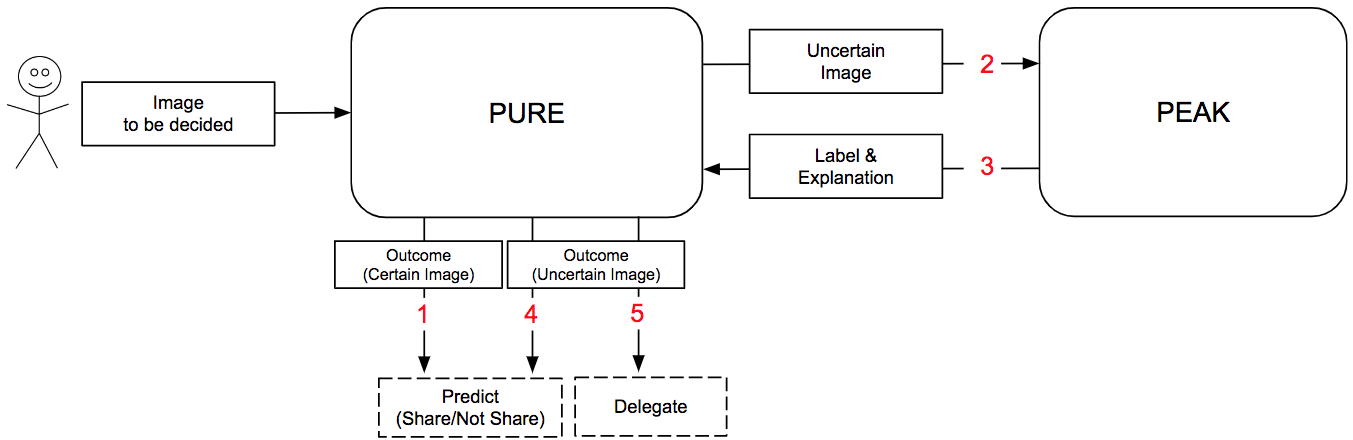}
    \caption{System overview schema of \peak~in combination with \pure}
    \label{fig:system_overview_schema}
\end{figure}
%
In order to reduce the number of images delegated to the user to decide on, \peak~is utilized for uncertain images, i.e.~images \pure~has difficulty making a prediction about.~\pure~only uses \peak's~privacy decisions for certain class-category pairs for which \peak~performs well.~For images where \peak~does not perform well, \pure~delegates the decision back to the user.~Hence, \peak~was not able to provide certainty for that images.~Figure \ref{fig:system_overview_schema} shows the schema of \peak~in combination with \pure.~A user gives an image to her personal privacy assistant (\pure), which then reports its privacy prediction (i.e.~share or not share) for the given image when certain about its prediction $\left ( 1 \right )$.~When not certain, \pure~first delegates the (uncertain) image to \peak~$\left ( 2 \right )$.~In this case, \peak~classifies the uncertain image and subsequently shares the decision and the generated explanation with \pure~$\left ( 3 \right )$.~After receiving input from \peak, \pure~finalizes its decision whether to use \peak's prediction result (i.e.~share or not share) $\left ( 4 \right )$ or to delegate the privacy decision to the user $\left ( 5 \right )$. 

When the uncertainty threshold, denoted as $\theta$, is set to $0.7$ by a user, $66\%$ of the images can be classified with \pure, i.e.~these are certain images and accuracy of $96\%$ is achieved~\cite{ayci2023uncertainty}. The remaining $34\%$ of images, a total of $1695$, are all uncertain images and these are delegated to the user by \pure. Hence, if \peak~is able to improve on this percentage and reduce the number of images delegated to the user whilst at the same time maintaining similar accuracy, it avoids the cognitive overload in the user and successfully improves upon \pure.

\begin{table}[htb]
\centering
\resizebox{\columnwidth}{!}{%
\begin{tabular}{|c|cc|cc|cc|cc|}
\hline
\multirow{2}{*}{}  & \multicolumn{2}{c|}{\textbf{Dominant}}                  & \multicolumn{2}{c|}{\textbf{Opposing}}                  & \multicolumn{2}{c|}{\textbf{Collaborative}}             & \multicolumn{2}{c|}{\textbf{Weak}}                      \\ \cline{2-9} 
                   & \multicolumn{1}{c|}{\textbf{public}} & \textbf{private} & \multicolumn{1}{c|}{\textbf{public}} & \textbf{private} & \multicolumn{1}{c|}{\textbf{public}} & \textbf{private} & \multicolumn{1}{c|}{\textbf{public}} & \textbf{private} \\ \hline
\textbf{All}       & 
\multicolumn{1}{c|}{5} & 8 & 
\multicolumn{1}{c|}{9} & 6 & 
\multicolumn{1}{c|}{30} & 32 & 
\multicolumn{1}{c|}{5} & 4 \\ \hline
\textbf{Uncertain} & 
\multicolumn{1}{c|}{3} & 11 & 
\multicolumn{1}{c|}{13} & 12 & 
\multicolumn{1}{c|}{11} & 37 & 
\multicolumn{1}{c|}{7} & 6 \\ \hline
\end{tabular}}
\caption{Percentage (\%) of all and uncertain images in the Training set that belong to the explanation categories}
\label{tab:training_percentages}
\end{table}

Table \ref{tab:training_percentages} shows distribution of all and uncertain images in the Training set across explanation categories. Firstly, it's clear the Collaborative category is the most important category since it comprises more than half of all images ($62\%$) whereas the other categories are featured in roughly similar frequencies. In terms of privacy class, there is no clear bias towards either public or private images with the difference at most $3\%$ within each category. Looking at only uncertain images, the Collaborative category is again the most common at $48\%$ of uncertain images, but there are much fewer public images compared to all images ($11\%$ versus $30\%$) and somewhat more private images. A second difference for uncertain images is that there are more images in the Opposing category ($25\%$ compared to $15\%$). For the Dominant and Weak categories, there are only limited differences between all and only uncertain images. Finally, in contrast to all images, uncertain images more frequently belong to the private class ($66\%$ compared to $50\%$).

\begin{table}[htb]
\centering
\resizebox{\columnwidth}{!}{%
\begin{tabular}{|c|cc|cc|cc|cc|}
\hline
\multirow{2}{*}{}  & \multicolumn{2}{c|}{\textbf{Dominant}}                  & \multicolumn{2}{c|}{\textbf{Opposing}}                  & \multicolumn{2}{c|}{\textbf{Collaborative}}             & \multicolumn{2}{c|}{\textbf{Weak}}                      \\ \cline{2-9} 
& \multicolumn{1}{c|}{\textbf{public}} & \textbf{private} & \multicolumn{1}{c|}{\textbf{public}} & \textbf{private} & \multicolumn{1}{c|}{\textbf{public}} & \textbf{private} & \multicolumn{1}{c|}{\textbf{public}} & \textbf{private} \\ \hline
\textbf{All}       & 
\multicolumn{1}{c|}{86} & \textbf{88} & 
\multicolumn{1}{c|}{90} & 56 & 
\multicolumn{1}{c|}{98} & \textbf{93} & 
\multicolumn{1}{c|}{80} & 72 \\ \hline
\textbf{Uncertain} & 
\multicolumn{1}{c|}{54} & \textbf{86} & 
\multicolumn{1}{c|}{85} & 55 & 
\multicolumn{1}{c|}{91} & \textbf{91} & 
\multicolumn{1}{c|}{70} & 73 \\ \hline
\end{tabular}}
\caption{Explanation category and class-specific privacy prediction performance (\% accuracy) of \peak~on all and uncertain images in the Training set}
\label{tab:training_performance}
\end{table}

\pure~decides whether to use the output of \peak~to make predictions or to delegate decisions for uncertain images back to the user based on two criteria, namely \textit{high} and \textit{consistent} performance of \peak. Hence, \peak~should perform well, i.e.~accuracy should be greater than $0.85$, and model performance on both all and uncertain images should be consistent, i.e.~the difference in accuracy between the two groups should be less than $5\%$. Table \ref{tab:training_performance} shows the privacy prediction performance of \peak~in terms of explanation category and class pairs. Based on the defined criteria, there are two \textit{Category-class} pairs with high and consistent performance, namely \textit{Dominant-private} and \textit{Collaborative-private}. Hence, for these images, PURE will use the output of \peak~to make a privacy prediction. 

The usefulness of \peak~should ultimately be determined based on performance for the uncertain images in the Test data. In the Test set, Dominant-private and Collaborative-private images comprise $7\%$ and $25\%$ respectively, capturing nearly one-third of all uncertain images. Additionally, \peak~performs well as prediction accuracy is $94\%$ and $93\%$ for Dominant-private and Collaborative-private, respectively. Hence, high accuracy that's broadly in line with the observed performance for the Training set. Overall, for nearly a third of the uncertain images, \peak~yields high performance. 

When \pure~delegates the selected uncertain images to \peak, fewer images are delegated to the user whilst classification accuracy is not compromised. Based on the Test set, the number of images delegated to the user is $23\%$, whereas it is $34\%$ for \pure~alone. Hence, it avoids cognitive overload in the user that user input is no longer needed for $542$ uncertain images. It reduces the risk of errors and biases by asking the user less, so a strong improvement due to \peak. Moreover, given the high accuracy for Dominant-private and Collaborative-private images, overall model performance is high at $0.959$, and nearly equal to \pure's standalone accuracy of $0.963$. Consequently, combining \pure~with \peak~results in nearly equal performance but for $77\%$ of the data as opposed to $66\%$ when it is only \pure. 

Finally, it's worth adding that in general, making a prediction for private images is by itself already challenging. However, \peak~is able to make accurate predictions in an even more challenging situation, namely uncertain private images. In doing so, \peak~is effective in reducing the number of decisions delegated to the user whilst at the same time achieving convincing privacy prediction performance. Hence, in view of these results, we answer RQ2 positively, namely \peak's generated explanations can indeed be used to improve personal assistants' decision-making.

\section{Conclusion}\label{sec-conclusion}
In this paper, we propose a novel privacy assistant (\peak) to understand why a given image is considered public or private based on the characteristics of the image.~Our privacy assistant is able to discover $20$ latent topics from descriptive image tags using topic modelling and subsequently makes privacy predictions based on the relationship between images and their associated topics.~We restrict the number of topics conservatively to simplify explanations while keeping predictions as accurate as possible using these topics.~We manually map each topic to a single textual representation (i.e.~topic name) in this paper only for presentation purposes.~Each topic is a proxy for a set of visual contexts that are referred to by a set of tags composing the topic.~This simple construct makes topics very flexible due to their limited semantic constraints, unlike a single textual representation of each topic.~However, users can represent each topic using a single textual representation as we do in this paper if it is more convenient for their consumption.~Moreover, our privacy assistant automatically generates explanations for privacy decisions.~The privacy classifier achieves high accuracy, demonstrating the effectiveness of the topic-based representation of images.~Furthermore, a user study shows the generated explanations are found to be sufficiently detailed, satisfying, and understandable.~Finally, our results show that \peak~can improve the decision-making of personal assistants by reducing the number of images delegated to the user whilst not compromising on model performance. 

An important direction for future work is to be able to get feedback from users and update the explanations.~In future work, we will deep dive into the Weak explanation category by clustering.~For instance, we can discover groups in the category by utilizing k-means or hierarchical clustering algorithms~\cite{likas2003global, johnson1967hierarchical}. We can automatically generate tags using different models/tools as well as Clarifai. For instance, multi-modal deep learning models such as CLIP~\cite{radford2021learning} can predict which text snippets (or tags) are related to a given image (and vice versa) using distances in the joint embedding space of text and images.~Given public and private images, we can extract all relevant words in the proximity of the image embeddings and consider the most discriminative words as tags with tag diversity in mind.~Moreover, the generated topics can be named manually or automatically.~Various automated methods are available for this purpose, such as selecting the most similar word to the tags as the topic name.~The similarity can be calculated using word embeddings of the tags or by utilizing an ontology like WordNet~\cite{wordnet}.~We leave the task of automatically naming the generated topics as future work.~As an interesting line of future research, \peak's explanation categories could be adapted to the Schwartz theory of basic values~\cite{schwartz2012overview}. Another future direction would be expanding our methodology to the classification of confidential documents since our approach is flexible and can work with text-based inputs as well.

\section{Acknowledgments}
This paper significantly extends~\cite{ayci-2023-ic,ayci-2023-aamas}, which present a general overview of the ideas.~The first author is supported by the Scientific and Technological Research Council of Turkey (TÜBİTAK) and Turkish Directorate of Strategy and Budget under the TAM Project number $2007K12-873$.~This research was partially funded by the Hybrid Intelligence Center, a $10$-year programme funded by the Dutch Ministry of Education, Culture and Science through the Netherlands Organisation for Scientific Research, \url{https://hybrid-intelligence-centre.nl}.

\section{Note to Reviewers}
This section is not meant to be part of the paper.~It describes differences between this paper and previously published work~\cite{ayci-2023-aamas,ayci-2023-ic}.~\cite{ayci-2023-aamas} is a 2-page extended abstract that mostly discusses the importance on privacy explanations.~\cite{ayci-2023-ic} is a 5-page magazine article that outlines the methodology and shows two example explanations that our approach could generate, while discussing challenges.~This paper provides the details as well as results that have not appeared in those works.~Only initial ideas from Section 3, 4.1, 4.2 and 5.1 have appeared before.~Sections 2, 4.3, and 5.2 onward (including the algorithm and all results) are new to this paper.

\bibliographystyle{ACM-Reference-Format}
\bibliography{references.bib}

\end{document}